\documentclass[lettersize,journal]{IEEEtran}
\usepackage{array}
\usepackage[caption=false,font=normalsize,labelfont=sf,textfont=sf]{subfig}
\usepackage{textcomp}
\usepackage{url}
\usepackage{verbatim}
\usepackage{cite}

\usepackage{xcolor}         
\usepackage{adjustbox}
\usepackage{amsmath}
\usepackage{amsthm}
\usepackage{amsfonts,amssymb} 
\usepackage{graphicx}
\usepackage{algorithm}
\usepackage{algorithmic}
\usepackage{setspace}
\usepackage{multirow}
\usepackage{pifont}
\usepackage{bbding}
\usepackage{color}
\usepackage{wrapfig}
\usepackage{bm}
\usepackage{colortbl}

\usepackage{booktabs}

\newcommand{\C}     {\mathcal{C}}

\newcommand{\N}     {\mathcal{N}}

\newcommand{\Ndis}     {\mathcal{N}(\bm{0}, \mathbf{I} )}

\newcommand{\y}     {\bm{y}}
\newcommand{\Y}     {\mathcal{Y}}
\newcommand{\f}     {\bm{f}}

\newcommand{\p}     {\bm{p}}
\newcommand{\q}     {\bm{q}}

\newcommand{\ep}     {\bm{\epsilon}}
\newcommand{\x}     {\bm{x}}
\newcommand{\X}     {\mathcal{X}}
\newcommand{\T}     {\mathcal{T}}

\newcommand{\z}     {\bm{z}}
\newcommand{\n}     {\bm{n}}

\newcommand{\rev}[1]{\textcolor{black}{{#1}}}

\DeclareMathOperator*{\argmin}{argmin}

\newcommand{\bihan}[1]{\textcolor{black}{#1}}
\newcommand{\siyuan}[1]{{\color{black}#1}}
\newcommand{\chong}[1]{\textcolor{black}{#1}}

\def\etal{\emph{et al.}}

\def\eg{\emph{e.g., }}

\begin{document}

\title{Taming Noise Initialization for Zero-shot Video Restoration with Image Diffusion Priors}



\author{Chong Wang, Zixuan Fu, Hao Cheng~\IEEEmembership{Member,~IEEE}, Siyuan Yang~\IEEEmembership{Member,~IEEE},\\ Alex C. Kot,~\IEEEmembership{Life Fellow,~IEEE}, Bihan Wen,~\IEEEmembership{Senior Member,~IEEE}
\thanks{
Chong Wang, Zixuan Fu, and Bihan Wen are with the School of Electrical and Electronic Engineering, Nanyang Technological University, Singapore 
(e-mail: \{wang1711, zixuan.fu, bihan.wen\}@ntu.edu.sg).

Hao Cheng is with the School of Artificial Intelligence, Hebei University of Technology, Tianjin, China (e-mail: chenghao@hebut.edu.cn)

Siyuan Yang is with the Division of Robotics, Perception and Learning, KTH Royal Institute of Technology, Stockholm, Sweden (email: siyuany@kth.se).

Alex C. Kot is with the Faculty of Engineering , Shenzhen MSU-BIT University, China.
(e-mail: eackot@ntu.edu.sg)

Corresponding author: Bihan Wen.
}
}

\markboth{Journal of \LaTeX\ Class Files,~Vol.~14, No.~8, August~2021}%
{Shell \MakeLowercase{\textit{et al.}}: A Sample Article Using IEEEtran.cls for IEEE Journals}


\maketitle

\begin{abstract}
Recent advances in diffusion models have achieved remarkable success in restoring high-quality images in a zero-shot manner.
\bihan{Unfortunately, extending this success to video restoration is challenging due to the substantial computational resources required to pre-train video diffusion models.}
A rising trend seeks to address video restoration by utilizing a pre-trained image diffusion model, initiating restoration from either frame-wise independent or identical Gaussian noise.
\bihan{Yet, these approaches often fall short, as such simple noise initiations fail to capture and utilize the inherent temporal correlations within degraded video frames.}
\chong{In this work, we propose taming noise initialization for zero-shot video restoration, }
\bihan{which preserves the intrinsic temporal correlations by generating initial noises directly using DDIM inversion on the corrupted video frames themselves.}
\bihan{Building on this, we introduce an adaptive noise blending strategy that suppresses degradation patterns encoded during DDIM inversion while gradually steering the noise toward a Gaussian distribution.}
To further emphasize the temporal consistency, we propose measurement-guided trajectory attention to better align features across frames.
\siyuan{Extensive experiments on multiple video restoration tasks demonstrate that our method outperforms existing zero-shot approaches, consistently achieving superior results on the DAVIS, REDS4, and Set8 datasets.}

\end{abstract}

\begin{IEEEkeywords}
Diffusion models, video restoration, zero-shot restoration.
\end{IEEEkeywords}

\begin{figure} [t]
  \centering 
    \includegraphics[width=0.95\linewidth]{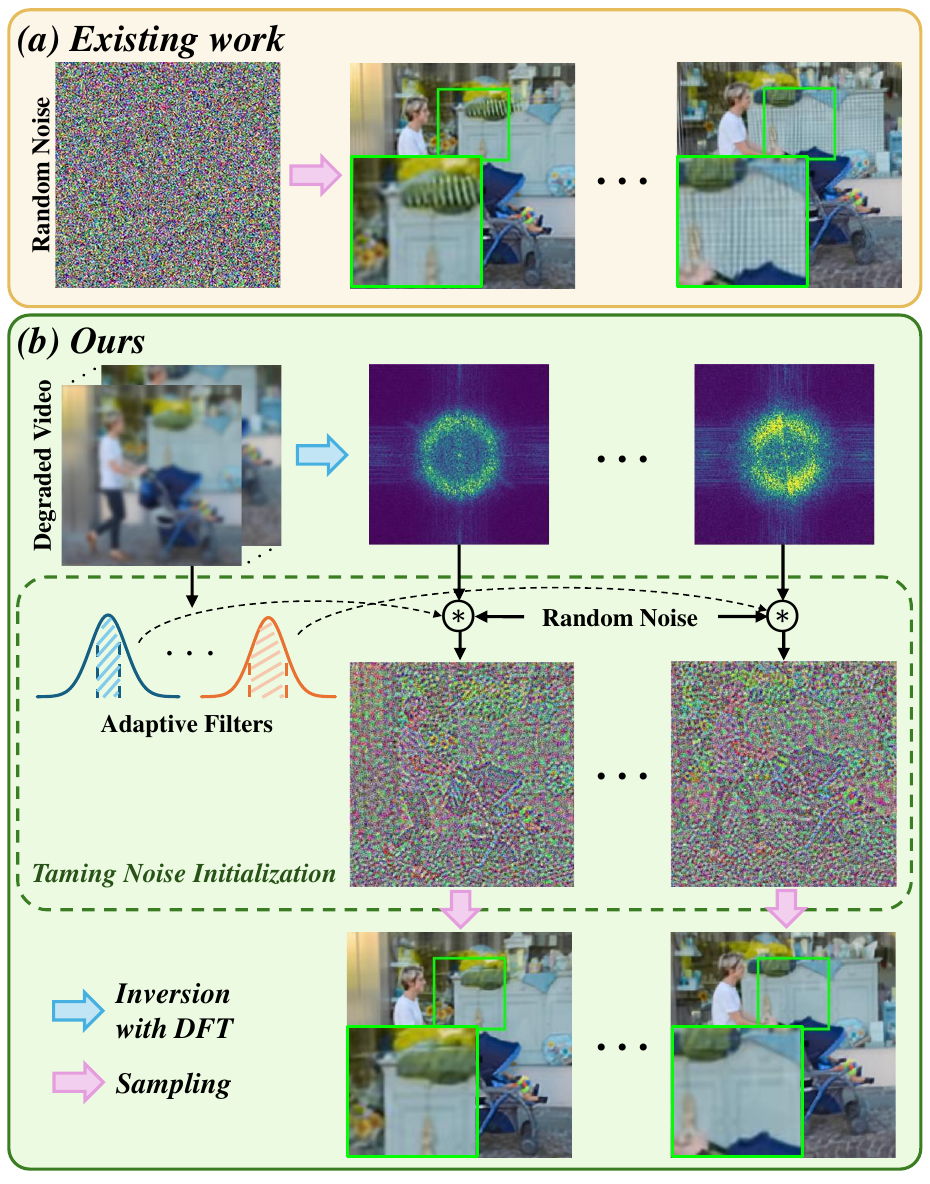}
\caption{(a) Existing method~\cite{kwon2024solving} utilizes a simple identical Gaussian noise initialization but fails to \siyuan{ensure frame consistency.}
%
\chong{(b) Our method tames the DDIM inversion noise with adaptive filtering while capturing the inherent temporal correlation in the degraded video, achieving results with guaranteed temporal consistency.}
}
\vspace{-3.5mm}
    \label{fig_intro}
  \end{figure} 

\section{Introduction}
\label{sec:intro}
\IEEEPARstart{D}{enoising} diffusion models~\cite{ho2020denoising}
have demonstrated exceptional performance in synthesizing images and videos with promising perceptual quality~\cite{ho2022cascaded, rombach2022high, xing2023survey, harvey2022flexible, ho2022imagen, jiang2024animediff, xu2024sgdm}.
Recently, such effectiveness in modeling complex distributions has motivated a wide application in both image and video-based restoration tasks~\cite{gao2023implicit, ren2023multiscale, guo2023boundary, guo2023shadowdiffusion, lugmayr2022repaint, zhou2024upscale, yang2025motion, zhang2024mmginpainting}.
%
\siyuan{While task-specific diffusion models achieve impressive results on their targeted problems, they frequently struggle with generalizability across various restoration tasks.}
Instead, another line of research tends to exploit the generative power inherent in a pre-trained diffusion network to address various image restoration problems in a zero-shot manner~\cite{zhu2023denoising, chung2023diffusion, chung2022improving, wang2022zero, song2024solving}.

\siyuan{Despite the success of zero-shot approaches in image restoration, extending these achievements to video restoration poses significant challenges. 
Pre-training video diffusion models from scratch requires substantial computational resources, making it impractical for various applications.}
%
%
\siyuan{Consequently,} how to effectively address zero-shot video restoration remains largely underexplored.
Very recently, \siyuan{researchers have attempted to bridge this gap by adapting pre-trained image diffusion models for video restoration without retraining~\cite{kwon2024solving, cao2024zero}.}
%
%
In~\cite{cao2024zero}, the authors proposed to utilize the temporal consistency guidance to warp the information between frames and prevent temporal misalignment in the restored video.
\siyuan{Alternatively,} another work~\cite{kwon2024solving} adopts a simple strategy that \siyuan{enforces identical initial noise across frames,}
thereby ensuring temporal consistency.
%
%


Video, known for its unique characteristics, contains time-correlated scenes and exhibits varying motion dynamics. 
However, existing approaches directly start restoration with a simple noise initiation, completely neglecting the temporal correlation across frames.
Specifically, \cite{kwon2024solving} \siyuan{ initializes restoration with identical noise for all frames,}
imposes \siyuan{an overly} strong restriction between frames, potentially constraining the \siyuan{natural} dynamics of the restored frames.
Conversely, \cite{cao2024zero} 
\siyuan{adopts} frame-wise completely independent Gaussian noises that may allow excessive freedom for each frame, thereby violating the temporal correlation across frames.
Therefore, determining how to appropriately impose \siyuan{temporal} correlations 
\siyuan{during noise initialization} is crucial for achieving promising \siyuan{video} restoration results.
\rev{To be more specific, the initialization determines the starting point of the sampling trajectory and strongly affects (i) how well the reverse process stays within the diffusion model learned data manifold, and (ii) how effectively the subsequent steps can correct the trajectory to satisfy the physical degradation model.}
To this end, we argue that the degraded videos 
\siyuan{inherently contain} sufficient temporally correlated information, which can be effectively utilized to initialize the ``right" noise for the diffusion restoration process.
%


In this work, we propose a novel zero-shot framework for video restoration, called \chong{\textit{TINDER}}, which only necessitates a pre-trained image diffusion prior.
Motivated by the success of DDIM inversion in video editing~\cite{ceylan2023pix2video, qi2023fatezero, geyer2023tokenflow, wang2024cove}, which establish a deterministic mapping between clean image and the corresponding noise sample.
\siyuan{Our method starts by performing DDIM inversion on the degraded video frames to obtain temporally correlated initial noise, effectively preserving temporal correlations and providing essential temporal guidance for the restoration process.}
\chong{Unfortunately, directly applying DDIM inversion in video restoration will produce sub-optimal results since the degradation pattern in the input frames can also be encoded into the inverted noise.}
%
%
%
\chong{\textbf{To this end, the main contribution of this paper is introducing an adaptive filtering strategy to tame noise initialization for video restoration.}}
%
\chong{Since video sequences inherently exhibit dynamic content, with significant variations in frequency distribution even across consecutive frames. 
Thus, our filtering strategy dynamically tailors filters based on the energy distribution of each frame, which is further utilized to blend the DDIM inversion results with Gaussian noise.}
%
%
This strategy not only enables varying yet adaptive filtering on each \rev{DDIM inverted noise}, but also compensates for the missing high-frequency energy caused by degradation \rev{in the inverted noise}.
\rev{Consequently, it steers the DDIM inversion noise towards a Gaussian distribution that better aligns with the diffusion process.
}
%
To further emphasize temporal consistency, we propose a measurement-guided trajectory attention during the diffusion sampling process.
More in detail,
\siyuan{motion trajectories are estimated from adjacent input frames and incorporated into the attention layer of the diffusion network, ensuring temporal consistency between consecutive frames.}
%
Finally, the proposed method offers a flexible framework for video restoration, where data consistency can be guaranteed by plugging any existing image-based diffusion sampling strategy.
\siyuan{Extensive experiments show that the proposed method consistently achieves superior performances in various video restoration tasks on the DAVIS, REDS4, and Set8 datasets.}
%

The main contributions of this work are as follows:
\begin{itemize}
    \item We propose \chong{\textit{TINDER}}, a novel zero-shot video restoration framework with image diffusion priors, where the initial noise is obtained via DDIM inversion to maintain the temporal relation in the degraded video frames.
    \item We propose an adaptive noise blending strategy to eliminate the degradation pattern in inversion noise for better restoration, followed by a measurement-guided trajectory attention to further guarantee the temporal consistency.
    \item \siyuan{Extensive experiments on multiple video restoration tasks demonstrate our methods outperform existing competing approaches on the DAVIS, REDS4, and Set8 datasets.}
\end{itemize}

\section{Related Work}
\label{sec:related_work}

\subsection{Conditional Diffusion for Video Restoration}
Denoising diffusion probabilistic models~\cite{ho2020denoising, sohl2015deep,nichol2021improved, ho2022cascaded, rombach2022high, dhariwal2021diffusion, ho2022video} have been proven to be a powerful class of unconditional generative models for high-quality image synthesis.
This iterative denoising mechanism not only results in high-fidelity and diverse image synthesis but also establishes a flexible framework that can be extended to various conditional tasks.
Building on their remarkable success in unconditional image generation, researchers have adapted these models for conditional settings by incorporating additional inputs such as degraded images.
This adaptation allows the model to generate outputs that are not only photorealistic but also context-specific, making them particularly well-suited for video restoration tasks where preserving spatial detail and temporal coherence is critical~\cite{zhou2024upscale,chen2024learning, yang2025motion}.
%
Traditional video restoration methods~\cite{wang2019edvr, zhou2022revisiting, kim2018spatio, cheng2021audio, chen2023self, baniya2023omnidirectional, ding2023blind, liang2024vrt, liang2022recurrent} typically employ deterministic mappings that 
\rev{learn a fixed transformation from degraded to restored frames.
Recent representative approaches employ Transformer architectures to capture long-range spatio-temporal dependencies while explicitly addressing motion and alignment across frames.
For instance, VRT~\cite{liang2024vrt} presents a Transformer-based framework with spatio-temporal attention and alignment mechanisms for high-quality video restoration, and RVRT~\cite{liang2022recurrent} further improves temporal modeling and efficiency via a recurrent formulation with guided deformable attention for correspondence-aware feature aggregation under complex motion.
Benefiting from task-specific supervision, these end-to-end methods typically achieve strong performance and fast feed-forward inference when the target degradation and data distribution match the training setting.}
In contrast, diffusion-based approaches embrace a probabilistic framework that models the full underlying distribution of clean video frames.
This probabilistic nature enables the model to capture intrinsic variability and uncertainty, leading to more robust restoration outcomes even in challenging conditions.
Several studies have explored the application of diffusion models in video restoration. 
For instance, Zhou et al.~\cite{zhou2024upscale} introduced Upscale-A-Video, a text-guided latent diffusion framework for video upscaling that ensures temporal coherence through integrated temporal layers and a flow-guided recurrent latent propagation module. 
Similarly, SATeCo~\cite{chen2024learning} leverages spatial-temporal guidance extracted from low-resolution videos to calibrate the high-resolution video denoising and restoration.

\subsection{Diffusion Models of Zero-shot Restoration}
Although these task-specific diffusion models excel in narrowly defined problems, they frequently struggle with generalizability across different restoration tasks.
Instead, another line of research tends to leverage pre-trained diffusion networks as generative priors to solve restoration tasks in a zero-shot manner~\cite{zhu2023denoising, chung2023diffusion, chung2022improving, wang2022zero, song2024solving, garber2024image}.

\noindent\textbf{Zero-shot Image Restoration.}
Existing zero-shot approaches typically apply constraints on the intermediate estimation, guiding the generative process to align with the degradation process of each specific image restoration task. 
For example, DDNM~\cite{wang2022zero} uses a range-null space decomposition on the clean image prediction, restricting diffusion to null-space components. 
A relaxed constraint is introduced in \cite{zhu2023denoising} by incorporating a least-squares penalty on the data term to ensure data consistency. 
In \cite{song2024solving}, a latent diffusion model enforces data consistency by optimizing the intermediate clean image prediction to minimize the data term during reverse sampling.
DPS~\cite{chung2023diffusion} further extends diffusion models to general nonlinear inverse problems by introducing a manifold-constrained gradient approximation, enabling wide applicability to noisy cases. 
Additionally, DDPG~\cite{garber2024image} proposes an iterative-preconditioned guidance approach that smoothly transitions from back-projection to least-squares gradient steps, enhancing noise robustness in image restoration tasks.

\noindent\textbf{Zero-shot Video Restoration.}
Building on the successes in image restoration, recent studies have explored extending these techniques to video data. 
Zero-shot video restoration presents unique challenges due to the need for maintaining temporal consistency across frames where simple frame-by-frame restoration methods often fail to address, leading to noticeable temporal flickering.
To overcome this, Cao et al.~\cite{cao2024zero} modify a pre-trained image diffusion model by incorporating cross-previous-frame attention layers. 
This adaptation explicitly models inter-frame dependencies, thereby mitigating flickering and ensuring that temporal features are coherently maintained across the sequence.
Yeh et al.~\cite{yeh2024diffir2vr} introduce hierarchical latent warping to leverage pre-trained image restoration diffusion models for video restoration. 
Kwon et al.~\cite{kwon2024solving} proposed a video inverse solver that utilizes image diffusion models, introducing a batch-consistent diffusion sampling strategy to ensure consistency across batches.
However, existing works only adopt a simple noise initiation, struggling to achieve temporal consistent results.
In this paper, we propose to initiate obtaining a temporal-correlated noise by exploiting the temporal correlation inherent in the degraded video, which provides crucial guidance for the restoration process.

\section{Preliminary}
\label{sec:preliminary}

\subsection{DDIM Inversion}

We adopt the most popular formulation of diffusion models as described in~\cite{ho2020denoising}, where there are two main processes.
The diffusion process begins with a clean image $\x_0 \sim p(\x)$ from a desired distribution, which is incrementally perturbed by adding small Gaussian noise $\ep \sim \Ndis$ as $\x_t = \sqrt{1-\beta_t} \x_{t-1} + \sqrt{\beta_t} \ep$, 
%
%
where $\{\beta_t\}_{t=1}^T$ defines a noise schedule in ascending order.
%
Through reparameterization, a perturbed sample $\x_t$ could be directly calculated from $\x_0$ in a closed form:
$\x_t = \sqrt{\bar{\alpha}_t} \x_0 + \sqrt{1-\bar{\alpha}_t} \ep$
where $\alpha_t = 1 - \beta_t$ and $\bar{\alpha}_t = \prod_{i=1}^t \alpha_i$.
The diffusion model $\ep_\theta(\cdot)$ is trained to approximate the reverse diffusion process, transitioning from a pure Gaussian noise $\x_T \sim \Ndis$ to a clean natural image $\x_0$.
The reverse process in~\cite{ho2020denoising} necessitates thousands of iterations, resulting in slow image generation.
As a remedy,~\cite{song2020denoising} proposed Denoising Diffusion Implicit Models (DDIM) which define a non-Markovian process as:
\begin{equation}\label{eq_ddim_sampling}
    \x_{t-1} = \sqrt{\alpha_{t-1}} \frac{\x_t - \sqrt{1-\alpha_t} \ep_\theta (\x_t)}{\sqrt{\alpha_t}} + \sqrt{1-\alpha_{t-1}} \ep_\theta (\x_t).
\end{equation}
As DDIM sampling in~\eqref{eq_ddim_sampling} is the deterministic process, it can be approximately reversed to map a clean image $x_0$ to a noise $x_T$ bijectively:
\begin{equation}\label{eq_ddim_inverse}
    \x_{t} = \sqrt{\alpha_{t}} \frac{\x_{t-1} - \sqrt{1-\alpha_{t-1}} \ep_\theta (\x_{t-1})}{\sqrt{\alpha_{t-1}}} + \sqrt{1-\alpha_{t}} \ep_\theta (\x_{t-1}).
\end{equation}
Thus we can establish a bijective link that bridges the noise with clean images.

\subsection{Diffusion Sampling with Data Consistency}\label{sec_sampling_with_dc}
In both image and video restoration, ensuring data consistency is essential for achieving reliable restoration results.
Existing methods~\cite{chung2023decomposed, zhu2023denoising, wang2022zero, chung2023diffusion} strive to apply constraints on the intermediate outputs of the diffusion sampling, iteratively guiding the restoration to align with the physics prior.
Notably, Chung \etal~\cite{chung2023decomposed} introduced a decomposed diffusion sampling approach that effectively integrates diffusion sampling with Krylov subspace methods.
They demonstrate that if the tangent space at a denoised sample $\x_{0|t} = (\x_t - \sqrt{1-\bar{\alpha}_t}\ep_\theta(\x_t))/\sqrt{\bar{\alpha}_t}$ constitutes a Krylov subspace, data consistency can be guaranteed by Conjugate Gradient method to solve the optimization as
\begin{equation}\label{eq_dds}
     \hat{\x}_{0|t} =  \argmin_{\x \in \x_{0|t} + \mathcal{K}_l} \lVert \y - A\x \rVert^2,
\end{equation}
where $\y$ is the degraded measurement, $A$ denotes the degradation process and $\mathcal{K}_l$ is the $l$-dimension Kyrlov subspace.
Inspired by~\cite{chung2023decomposed}, such strategy has been extended to video restoration in~\cite{kwon2024solving}.
Another research~\cite{zhu2023denoising} incorporates data consistency by utilizing a weighted sum of least-squares as
\begin{equation} \label{eq_diffpir}
    \hat{\x}_{0|t} = \argmin_{\x} \lVert \y - A\x \rVert^2 + \rho \lVert \x - \x_{0|t}  \rVert^2,
\end{equation}
where $\rho$ is the weighting parameter.

\begin{figure*} [t]
  \centering 
    \includegraphics[width=0.93\linewidth]{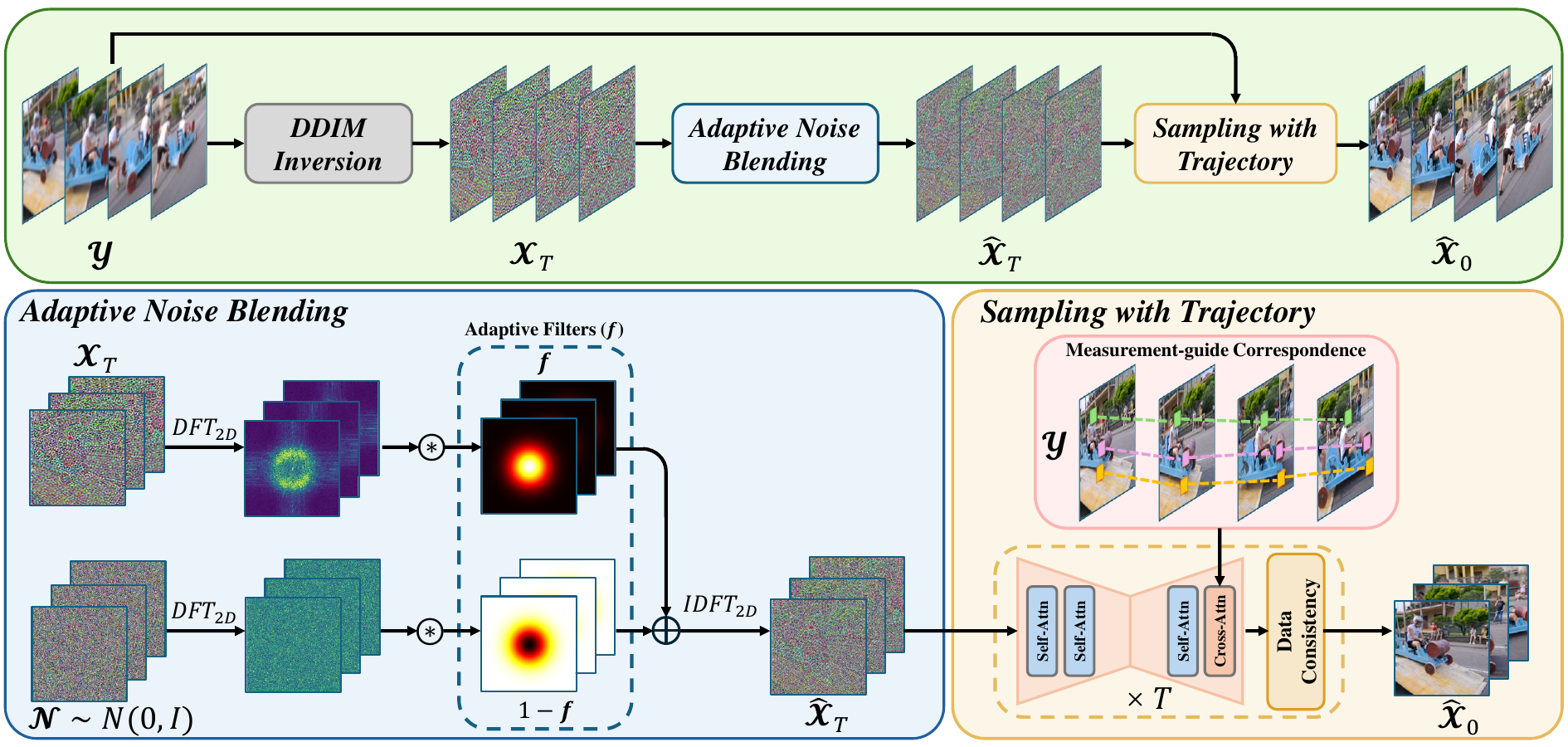}
\caption{Overview of the proposed \chong{\textit{TINDER}} framework, where the initial noise is first obtained by performing DDIM inversion on the degraded video to capture the inherent temporal correlation.
Subsequently, an adaptive noise blending strategy is employed to mitigate the degradation pattern in the inversion noise $\X_T$, while directing the noise toward a Gaussian distribution.
After that, the temporal correlation is further exploited by measurement-guided trajectory attention during the sampling process.
}
\vspace{-5mm}
    \label{fig_framework}
\end{figure*}

\section{Method}
\label{sec:method}
We start with an introduction to video restoration problem formulation.
Given a degraded video with $N$ frames $\Y = \{\y^i\}^{N}_{i=1}$, we aim to recover the underlying clean video $\X = \{\x^i\}^{N}_{i=1}$.
The degradation process can be modeled as performing degradation on each frame independently:
\begin{equation}
    \y^i = A \x^i + \n^i,
\end{equation}
where $A$ denotes the forward degradation process on each video frame, and $\n$ is the additive Gaussian noise.
\begin{figure} [t]
  \centering 
    \includegraphics[width=0.95\linewidth]{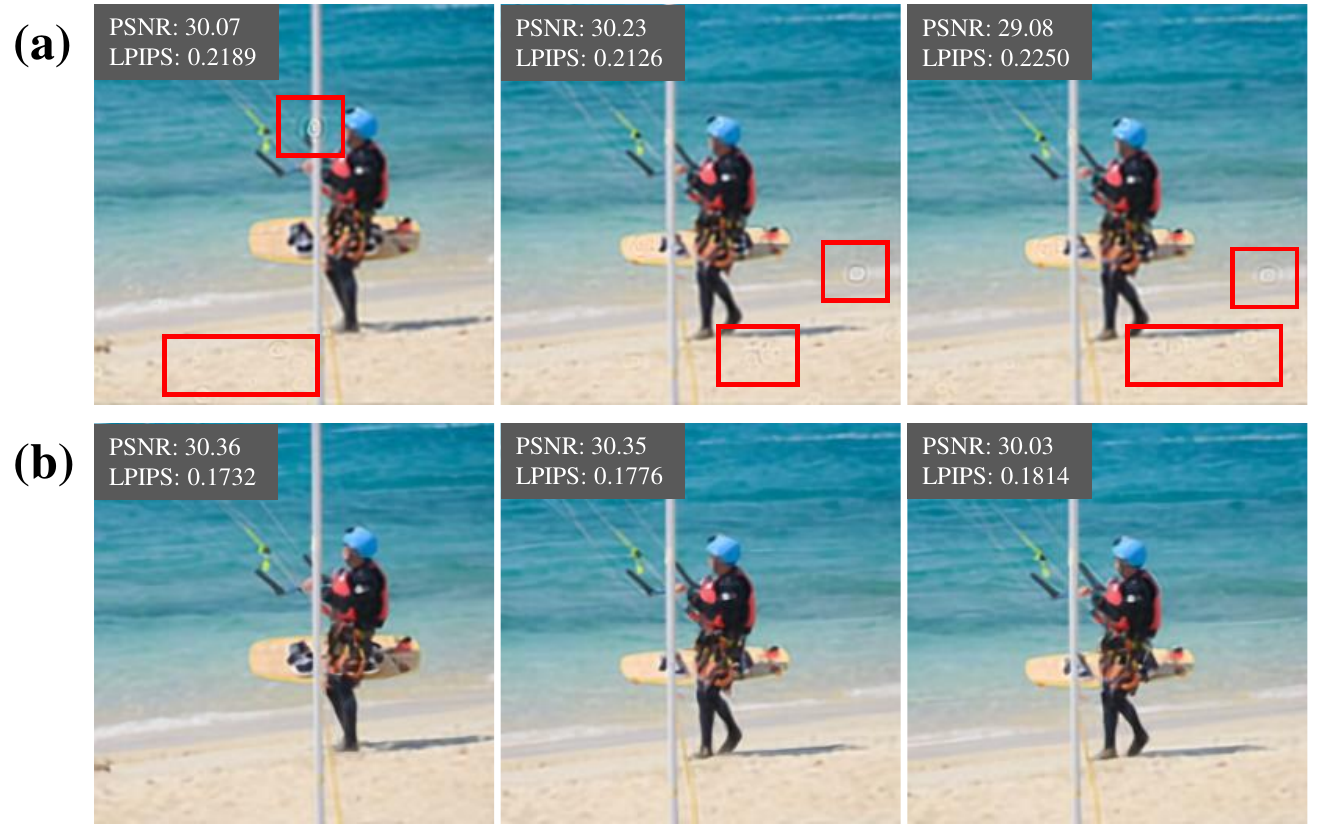}
\caption{Visual comparison of restoration using (a) DDIM inversion noise and (b) DDIM inversion with adaptive noise blending.
Directly using inversion noise for restoration encodes the degradation pattern into restored results, as shown in red boxes in (a).
}
    \label{fig_ddiminv}
    \vspace{-5mm}
\end{figure} 

\subsection{Overall Framework}
The proposed framework aims to solve the video restoration problem while only necessitating a pre-trained image diffusion model without re-tuning for each specific task.
\siyuan{As shown in Figure~\ref{fig_framework}, our framework can be divided into three main components: 1) DDIM inversion for initial noise generation, 2) adaptive noise blending for degradation pattern mitigation, and 3) sampling with trajectory attention for temporal consistency enhancement.}
In contrast to the existing works~\cite{kwon2024solving, cao2024zero} where the Gaussian noise is adopted as the starting point of diffusion sampling, we initiate the restoration process by obtaining temporal-correlated noises via performing DDIM inversion on the input video frames, denoted as $\X_T =\{ \x_T^i \}^N_{i=1} = \{\textit{DDIM}_{inv}(\y^i)\}^N_{i=1}$.
With the merit of DDIM inversion, the temporal correlation across each frame in the input video $\Y$ can be encoded into the initial inversion noise $\X_T$, thus offering effective guidance for the \siyuan{subsequent} restoration process.
However, as the initial noise $\X_T$ is \siyuan{derived} from the input degraded video, the undesired degradation pattern can also be preserved and encoded through the DDIM inversion.
Thus, directly performing diffusion sampling starting from $\X_T$ could cause significant artifacts and hinder the restored results, as shown in Figure~\ref{fig_ddiminv}.
To address this issue, we propose an adaptive noise blending strategy to mitigate the encoded degradation in initial inversion noise (Section~\ref{sec:noise_blending}).
Lastly, we further propose a measurement-guided trajectory attention mechanism (Section~\ref{sec:trajectory attention}) to achieve temporal
consistent restoration during diffusion sampling.

\subsection{Adaptive Noise Blending}\label{sec:noise_blending}
\begin{figure} [t]
  \centering 
    \includegraphics[width=0.95\linewidth]{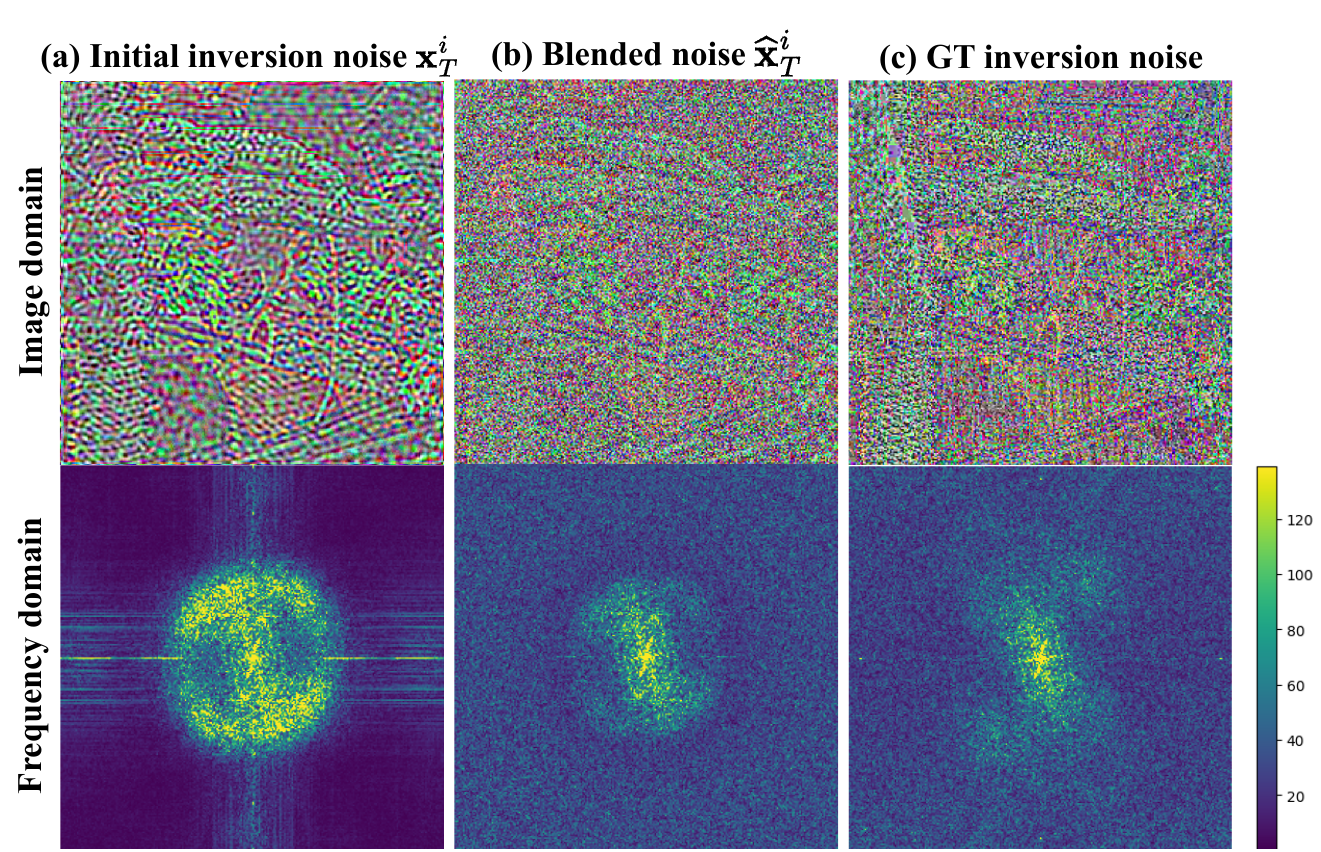}
\caption{A visual example of (a) the initial inversion noise obtained directly from the degraded frame, which exhibits significant degradation patterns and distorted frequency components.
With adaptive noise blending, (b) the blended noise increasingly resembles (c) the ground truth inverted noise, both in the image domain and in terms of energy distribution across the frequency domain.}
    \label{fig_noise_blend_visual}
    \vspace{-3mm}
\end{figure}

In the context of video restoration, the noise is initialized via performing DDIM inversion on the degraded video frames.
The inverted noise contains significant degradation patterns and directly adopting it for sampling will lead to sub-optimal restoration results.
Figure~\ref{fig_noise_blend_visual} (a) and (c) present the \siyuan{image} and frequency domain of the inverted noises using a degraded frame and corresponding clean frame, respectively.
The initial inversion noise contains noticeable jagged patterns with details being smeared out compared to the corresponding ground truth (GT) inversion noise.
This discrepancy is particularly pronounced in the frequency domain, where high frequencies are either completely lost or distorted, compelling the restoration process to infer or amplify the missing information.
We can notice that the low frequency components of initial inversion noise are relatively accurate compared to the high frequency.
Based on this observation, one may trivially adopt a pre-defined low-pass filter (LPF) to perform filtering on the initial inversion noise, which is common sense.
However, video is known for its dynamic content\siyuan{, with significant variations in frequency distribution even between frames of the same video.}
%
Therefore, directly \siyuan{applying a uniform pre-defined LPF across different frames would yield suboptimal results.}
%

To address this challenge, we introduce an adaptive noise blending strategy that tailors the LPF based on the energy distribution of each degraded frame.
Specifically, for a given a degraded frame $\x_T^i$, a Gaussian LPF $\f(c^i)$ is determined by optimizing the cutoff frequency $c^i$ that retains a fragment of energy of $\x_T^i$ as follows:
\begin{equation} \label{eq_get_filter}
    c^i = \argmin_{c} \left| \frac{\lVert DFT_{2D}(\x_T^i) * \f(c)\rVert^2_F}{\lVert DFT_{2D}(\x_T^i)\rVert^2_F} - \eta \right|,
\end{equation}
where $DFT_{2D}$ denotes the 2D discrete Fourier transform, $*$ is the Hadarmard product, and $\eta$ represents the proportion of retained energy.
This problem can be simply solved as $\lVert DFT_{2D}(\x_T^i) * \f(c)\rVert^2_F$ monotonically increasing with $c$.
Consequently, we can progressively increase the value of $c$ and halt the process once $c$ exceeds the energy coefficient $\eta$.
Once the LPF is optimized according to each frame $i$, the initial inversion noise is blended with a Gaussian noise as:
\begin{equation}
    \hat{\x}^i_T = \f(c^i) * \x^i_T + (\mathbf{1}- \f(c^i)) * \ep^i.
\end{equation}
Figure~\ref{fig_noise_blend_visual} (b) provides a visualization of the blended inversion noise, demonstrating greater similarity to the ground truth inversion noise, particularly in terms of energy distribution in the frequency domain.

\begin{table*}[t]
    \centering
    \caption{
    Quantitative evaluation results (PSNR, SSIM, LPIPS, and FVD) of video restoration tasks: \textit{Gaussian} deblur, super-resolution (2$\times$ and 4$\times$) and 70$\%$ inpainting on DAVIS~\cite{pont20172017}, REDS4~\cite{nah2019ntire}, and Set8~\cite{tassano2020fastdvdnet}, respectively.
    The best results are marked in \textbf{Bold}.}
    \setlength{\tabcolsep}{0.5mm}
    \adjustbox{width=0.99\linewidth}{
    \begin{tabular}{clcccccccccccc}
    \toprule
          &&\multicolumn{4}{c}{\textbf{DAVIS}~\cite{pont20172017}}&\multicolumn{4}{c}{\textbf{REDS4}~\cite{nah2019ntire}}& \multicolumn{4}{c}{\textbf{Set8}~\cite{tassano2020fastdvdnet}}\\
      \cmidrule(lr){3-6} \cmidrule(lr){7-10} \cmidrule(lr){11-14} 
        &Method &PSNR$\uparrow$&SSIM$\uparrow$&LPIPS$\downarrow$&FVD$\downarrow$&PSNR$\uparrow$&SSIM$\uparrow$&LPIPS$\downarrow$&FVD$\downarrow$&PSNR$\uparrow$&SSIM$\uparrow$&LPIPS$\downarrow$&FVD$\downarrow$\\
        \midrule
         \multirow{5}*{\rotatebox{90}{\textbf{Deblur}}}&
         DDS~\cite{chung2023decomposed} + independent &27.93 &0.8198 &0.1870 &69.08 &25.29 &0.7789 &0.2218 &59.33 &27.06 &0.7949 &0.2102 &83.60 \\
         &DDS~\cite{chung2023decomposed} + mix~\cite{ge2023preserve} &27.99 &0.8226 &0.1862 &69.25 &25.27 &0.7790 &0.2215 &60.79&27.07 &0.7951 &0.2107 &83.52 \\
         &DDS~\cite{chung2023decomposed} + prog~\cite{ge2023preserve} &27.99 &0.8220 &0.1869 &70.38 &25.21 &0.7768 &0.2219 &62.49 &27.19 &0.8000 &0.2073 &82.92 \\
         &Kwon~\etal\cite{kwon2024solving} \footnotesize{(DDS+identical)} &27.90 &0.8185 &0.1873 & 72.26 &25.25 &0.7787 &0.2195 & 65.31 &26.91 &0.7849 &0.2129 &88.22 \\
         &\cellcolor{gray!20}Ours (DDS) &\cellcolor{gray!20}\textbf{29.43} &\cellcolor{gray!20}\textbf{0.8608} &\cellcolor{gray!20}\textbf{0.1763} &\cellcolor{gray!20}\textbf{63.42}&\cellcolor{gray!20}\textbf{27.53} &\cellcolor{gray!20}\textbf{0.8357} &\cellcolor{gray!20}\textbf{0.2142} &\cellcolor{gray!20}\textbf{58.85}&\cellcolor{gray!20}\textbf{29.10} &\cellcolor{gray!20}\textbf{0.8547} &\cellcolor{gray!20}\textbf{0.1909} &\cellcolor{gray!20}\textbf{81.06}\\ 

         \midrule
         \multirow{5}*{\rotatebox{90}{\textbf{SR (2$\times$)}}}
         &DDS~\cite{chung2023decomposed} + independent &29.74 &0.8790 &0.0996 &21.13 &26.66 &0.8456 &0.1265 &16.18 &29.45 &0.8670 &0.1045 &22.36 \\
         &DDS~\cite{chung2023decomposed} + mix~\cite{ge2023preserve} &29.75 &0.8788 &0.0994 &22.64 &26.64 &0.8454 &\textbf{0.1263} &17.63 &29.42 &0.8670 &0.1048 &21.45 \\
         &DDS~\cite{chung2023decomposed} + prog~\cite{ge2023preserve} &29.78 &0.8799 &\textbf{0.0990} &23.95  &26.62 &0.8446 &0.1269 &20.59 &29.47 &0.8682 &\textbf{0.1039} &27.25 \\
         &Kwon~\etal\cite{kwon2024solving} \footnotesize{(DDS+identical)} &29.76 &0.8789 &0.0998 &25.21 &26.64 &0.8444 &0.1274 &22.05 &29.46 &0.8688 &0.1045 &25.53 \\
         &\cellcolor{gray!20}Ours (DDS) &\cellcolor{gray!20}\textbf{31.92} &\cellcolor{gray!20}\textbf{0.9204} &\cellcolor{gray!20}0.1032 &\cellcolor{gray!20}\textbf{18.48} &\cellcolor{gray!20}\textbf{29.45} &\cellcolor{gray!20}\textbf{0.9044} &\cellcolor{gray!20}0.1363 &\cellcolor{gray!20}\textbf{14.29} &\cellcolor{gray!20}\textbf{31.87} &\cellcolor{gray!20}\textbf{0.9193} &\cellcolor{gray!20}0.1147 &\cellcolor{gray!20}\textbf{20.63}\\

         \midrule
         \multirow{5}*{\rotatebox{90}{\textbf{SR (4$\times$)}}}&
         DDS~\cite{chung2023decomposed} + independent &26.89 &0.7810 &0.2537 & 167.12 &24.61 &0.7383 &0.3076 &149.21 &26.07 &0.7612 &0.2914 &179.26\\
         &DDS~\cite{chung2023decomposed} + mix~\cite{ge2023preserve} &26.90 &0.7811 &0.2538 & 164.53 &24.62 &0.7386 &\textbf{0.3073} &144.76 &26.05 &0.7613 &\textbf{0.2911} &182.51 \\
         &DDS~\cite{chung2023decomposed} + prog~\cite{ge2023preserve} &26.88 &0.7806 &\textbf{0.2533} &160.62 &24.61 &0.7383 &0.3077 &138.26 &26.07 &0.7614 &0.2913 &\textbf{170.55} \\
         &Kwon~\etal\cite{kwon2024solving} \footnotesize{(DDS+identical)} &26.89 &0.7806 &0.2538 & \textbf{155.58} &24.62 &0.7384 &0.3079 &\textbf{137.58} &26.08 &0.7617 &0.2913 &176.50 \\
         &\cellcolor{gray!20}Ours (DDS) &\cellcolor{gray!20}\textbf{27.09} &\cellcolor{gray!20}\textbf{0.7872} &\cellcolor{gray!20}0.2701 &\cellcolor{gray!20} 258.18 &\cellcolor{gray!20}\textbf{24.91} &\cellcolor{gray!20}\textbf{0.7468} &\cellcolor{gray!20}0.3325 &\cellcolor{gray!20} 233.21&\cellcolor{gray!20}\textbf{26.27} &\cellcolor{gray!20}\textbf{0.7688} &\cellcolor{gray!20}0.3055 &\cellcolor{gray!20} 294.58\\ 

        \midrule
         \multirow{5}*{\rotatebox{90}{\textbf{Inpainting}}}&
         DDS~\cite{chung2023decomposed} + independent &25.86 &0.7806 &\textbf{0.2338} & 240.72 &23.77 &0.7473 &0.2557 &287.74 &24.34 &0.7388 &0.2794 &541.28 \\
         &DDS~\cite{chung2023decomposed} + mix~\cite{ge2023preserve} &25.85 &0.7804 &0.2339 & 257.64 &23.80 &0.7481 &0.2552 &293.65 &24.33 &0.7388 &0.2795 &552.04 \\
         &DDS~\cite{chung2023decomposed} + prog~\cite{ge2023preserve} &25.82 &0.7798 &0.2347 &206.10 &23.78 &0.7474 &0.2553 &205.76 &24.33 &0.7388 &0.2788 &322.22 \\
         &Kwon~\etal\cite{kwon2024solving} \footnotesize{(DDS+identical)} &25.85 &0.7799 &0.2348 & \textbf{191.36} &23.76 &0.7466 &0.2563 &\textbf{187.25} &24.34 &0.7389 &0.2791 &\textbf{258.08} \\
         &\cellcolor{gray!20}Ours (DDS) &\cellcolor{gray!20}\textbf{27.08} &\cellcolor{gray!20}\textbf{0.8075} &\cellcolor{gray!20}0.2349 &\cellcolor{gray!20}270.70 &\cellcolor{gray!20}\textbf{25.46} &\cellcolor{gray!20}\textbf{0.7880} &\cellcolor{gray!20}\textbf{0.2551} &\cellcolor{gray!20}267.00&\cellcolor{gray!20}\textbf{25.77} &\cellcolor{gray!20}\textbf{0.7743} &\cellcolor{gray!20}\textbf{0.2772} &\cellcolor{gray!20}530.15\\ 

      \bottomrule
    \end{tabular}
    }
    \vspace{-3mm}
    \label{tab_dds_exp}
\end{table*}

\noindent\textbf{Why it works?}
The proposed noise blending strategy aims to adaptively remove the degradation pattern encoded in the initial DDIM inversion noise and provide reliable information for the subsequent restoration process.
The merits of this strategy can be categorized into two aspects:
\begin{itemize}
    \item First, \chong{the adaptive noise blending strategy enables a dynamic filtering according to the energy distribution of each frame with a single energy parameter $\eta$.}
    \item \rev{Besides, this adaptive filtering mitigates artifacts in the inversion noise by preserving the relatively accurate low-frequency and restoring high-frequency components via Gaussian blending, which yields a more balanced initialization that better matches the diffusion prior.}
\end{itemize}

\begin{algorithm}[t]
\caption{\chong{\textit{TINDER}}} 
\label{alg1}
\begin{algorithmic}
\REQUIRE degraded video $\Y = \{\y^i\}^{N}_{i=1}$, pre-trained image diffusion model $\ep_\theta$, stochasticity strength parameter $\zeta$.\\
\textbf{Initialization:} 
$\bm{\C} = \{\C^i\}_{i=1}^{N-1} \gets \{\y^i, \y^{i+1}\}_{i=1}^{N-1}$ via \eqref{eq_get_corr}\\
\textit{\textcolor{gray}{\# Initial noise with DDIM inversion}}\\
$\{ \x_T^i \}^N_{i=1} = \textit{DDIM}_{inv}(\{\y^i\}^N_{i=1})$\\
\textit{\textcolor{gray}{\# Adaptive noise blending}}\\
$\f(c^i) \gets \x_T^i$ via \eqref{eq_get_filter} \\
$\ep^i \sim \N (\bm{0}, \bm{I})$\\
$\hat{\x}^i_T = \f(c^i) * \x^i_T + (\mathbf{1}- \f(c^i)) * \ep^i$\\
$\hat{\X}_T = \{\hat{\x}^i_T\}_{i=1}^N$\\
\textit{\textcolor{gray}{\# Perform diffusion sampling for restoration}}\\
\FOR{$t = T,\dots ,1$}
\STATE \textit{\textcolor{gray}{\# Tweedie denoising with trajectory guidance}}
\STATE $\hat{\X}_{0|t} = (\hat{\X}_t - \sqrt{1-\bar{\alpha}_t} \ep_\theta(\hat{\X}_t, \bm{\C}; t)) / \sqrt{\bar{\alpha}_t}$
\STATE \textit{\textcolor{gray}{\# Data consistency}}
\STATE Update $\hat{\X}_{0|t}$ with Eq. \eqref{eq_dds} or Eq. \eqref{eq_diffpir}
\STATE \textit{\textcolor{gray}{\# Renoising}}
\STATE $\ep_t \sim \Ndis$
\STATE $\hat{\X}_{t-1} = $
\STATE $\quad \sqrt{\bar{\alpha}_{t-1}} \hat{\X}_{0|t} + \sqrt{1 - \bar{\alpha}_{t-1} - \zeta^2} \ep_\theta(\hat{\X}_t, \bm{\C}; t) + \zeta \ep_t $
\ENDFOR
\STATE $\textbf{return} \;\hat{\X}_0$
\end{algorithmic}
\end{algorithm}

\subsection{Measurement-guided Trajectory Attention}
\label{sec:trajectory attention}
As pre-trained image diffusion models address the restoration on a frame-by-frame basis, it is crucial to impose temporal correlation during the restoration process.
To further guarantee temporal consistency,
we propose a measurement-guided trajectory attention (MGTA) mechanism that can be seamlessly integrated into pre-trained image diffusion models.
%
The process begins by extracting temporal information via estimating the correspondence maps between adjacent frames in the degraded video.
Given a pair of adjacent frames $\y^i$ and $\y^{i+1}$, let $\phi(\y^i[p^i])$ denotes the patch centered at the pixel $p^i$ in the frame $\y^i$, the corresponding matching point $q^{i+1}$ in the next frame $\y^{i+1}$ is searched as:
\begin{equation} \label{eq_get_corr}
    q^{i+1} = \argmin_q \lVert \phi(\y^{i+1}[q]) - \phi(\y^i[p^i]) \rVert^2_F.
\end{equation}
Therefore, the correspondence map $\C^i = \{\p^i, \q^{i+1}\}_{i=1}^{N-1}$ is compiled by collecting these matched points.
To utilize the correspondence map to guide the attention module, we downsampled it to the resolution of each latent space.\footnote{Here, we simplify notation by using $\C^i$ to refer to the correspondence maps across all latent spaces.}
Inspired by the previous works~\cite{wu2023tune, ceylan2023pix2video, qi2023fatezero}, we extend the spatial self-attention module in the diffusion model to the spatial-temporal domain.
Besides, temporal consistency is reinforced with the aid of correspondence map $\C^i$ as:
\begin{equation}
\begin{split}
    Q &= W^Q \z^i, \\
    K=W^K[\z^i, \T(\z^{i+1}; \C^i)]&, V = W^V [\z^i;\T(\z^{i+1}; \C^i)],\\
    \text{Attn}(Q,K,V) &= \text{Softmax}(\frac{QK^\top}{\sqrt{d}})V,
\end{split}
\end{equation}
where $[\cdot]$ denotes the concatenation, $\z^i$ is the feature of the $i$-th frame and $\T$ transform that leverages trajectory maps $\C^i$ to align the feature $\z^{i+1}$ with those of the current frame.
\siyuan{The latent features $\z$ are refined through MGTA to mitigate the adverse effects of aggregating irrelevant patches in dense spatial-temporal attention.}
%
Following~\cite{ceylan2023pix2video}, we perform MGTA in the decoder layers of the UNet during sampling.
Overall, our method provides a flexible framework for video restoration where the data consistency can be directly guaranteed by plug-in existing image-based diffusion sampling strategies as discussed in Section~\ref{sec_sampling_with_dc}.
The overall steps of our framework are described in Algorithm~\ref{alg1}.

\section{Experiments}
\label{sec:experiments}

\subsection{Experimental Setup}
\noindent\textbf{Datasets and tasks.} To evaluate the efficacy of the proposed method, we conducted comprehensive experiments across three video datasets: DAVIS~\cite{pont20172017}, REDS4~\cite{nah2019ntire}, and Set8~\cite{tassano2020fastdvdnet}. 
For each dataset, we sampled 100 16-frame video clips as the ground truth for evaluation.
To demonstrate the generalization capability of our method, we conduct three representative video restoration tasks including \textit{Gaussian} deblur, super-resolution (SR), and inpainting.
Following~\cite{kwon2024solving, zhu2023denoising}, the degradation is performed using 2$\times$ and 4$\times$ bicubic down-sampling for SR.
For \textit{Gaussian} deblurring, a blur kernel of sized 13$\times$13 with a standard deviation of 2.0 is employed.
For video inpainting, we randomly drop 70$\%$ pixels in each video frame.

\noindent\textbf{Metrics.} For video restoration tasks, evaluating both distortion and perceptual quality is crucial.
Thus, we employ the Peak Signal-to-Noise Ratio (PSNR) and Structural Similarity Index (SSIM) to assess the fidelity of the restored videos.
For perceptual quality, we use the Learned Perceptual Image Patch Similarity (LPIPS) distance~\cite{zhang2018unreasonable}.
Additionally, a video-specific metric, namely Fréchet Video Distance (FVD)~\cite{unterthiner2019fvd} is utilized for comprehensive evaluation.

\begin{table}[t]
    \centering
    \caption{Hyper-paramters setting for each task.}
    \adjustbox{width=0.95\linewidth}{
    \begin{tabular}{l|cc|ccc}
    \toprule
         \textbf{Datasets}&&&\multicolumn{1}{c}{\textbf{DAVIS}}& \multicolumn{1}{c}{\textbf{REDS4}}& \multicolumn{1}{c}{\textbf{Set8}}\\
     \midrule
    Task &$T_{inv}$&$T$&$\eta$&$\eta$&$\eta$\\
        \midrule
         Deblur (\textit{Gaussian})& 20 & 20 & 0.5 & 0.7 & 0.7\\
         SR (2$\times$)& 20 & 20 & 0.7 & 0.7 & 0.7 \\
         SR (4$\times$)& 20 & 50 & 0.1 & 0.1 & 0.1 \\
         Inpaint (10$\%$)& 20 & 50 & 0.1 & 0.1 & 0.1 \\
      \bottomrule
    \end{tabular}
    }
    \label{tab_hyperparameter}
\end{table}

\subsection{Implementation Details}
The proposed method is implemented on PyTorch using NVIDIA RTX A5000.  
The total iterations for the DDIM inversion are fixed at $T_{inv}=20$ for each task and implementation.
For tasks such as Gaussian deblurring and 2$\times$ SR, the sampling iteration step is set to $T=20$.
For the more challenging tasks of 4$\times$ SR and 70$\%$ inpainting, a higher iteration of $T=50$ is required. 

The proposed method and all competing baselines necessitate a pre-trained image diffusion as the prior.
To conduct a fair comparison, we adopt the same pre-trained diffusion model sourced from~\cite{dhariwal2021diffusion} for all experiments.
The hyper-parameter setting is listed in Table~\ref{tab_hyperparameter}.
%
%
The parameter $\eta$ controls the energy partition preserved during the adaptive noise blending.
For a task with less degradation severity, the $\eta$ can be relatively high, \eg around 0.7, indicating that we can preserve more energy in the DDIM inversion noise to maintain the correlation across each frame.
In contrast, for those tasks with severe degradation, \eg 4$\times$ SR and 70$\%$ inpainting, the $\eta$ should be relatively low ($\eta=0.1$) since the inverted noise contains significant degradation.
Therefore, a smaller $\eta$ indicating more aggressive filtering is required to remove the severe degradation pattern encoded in the inversion noises.

\begin{figure*} [t]
  \centering 
    \includegraphics[width=0.99\linewidth]{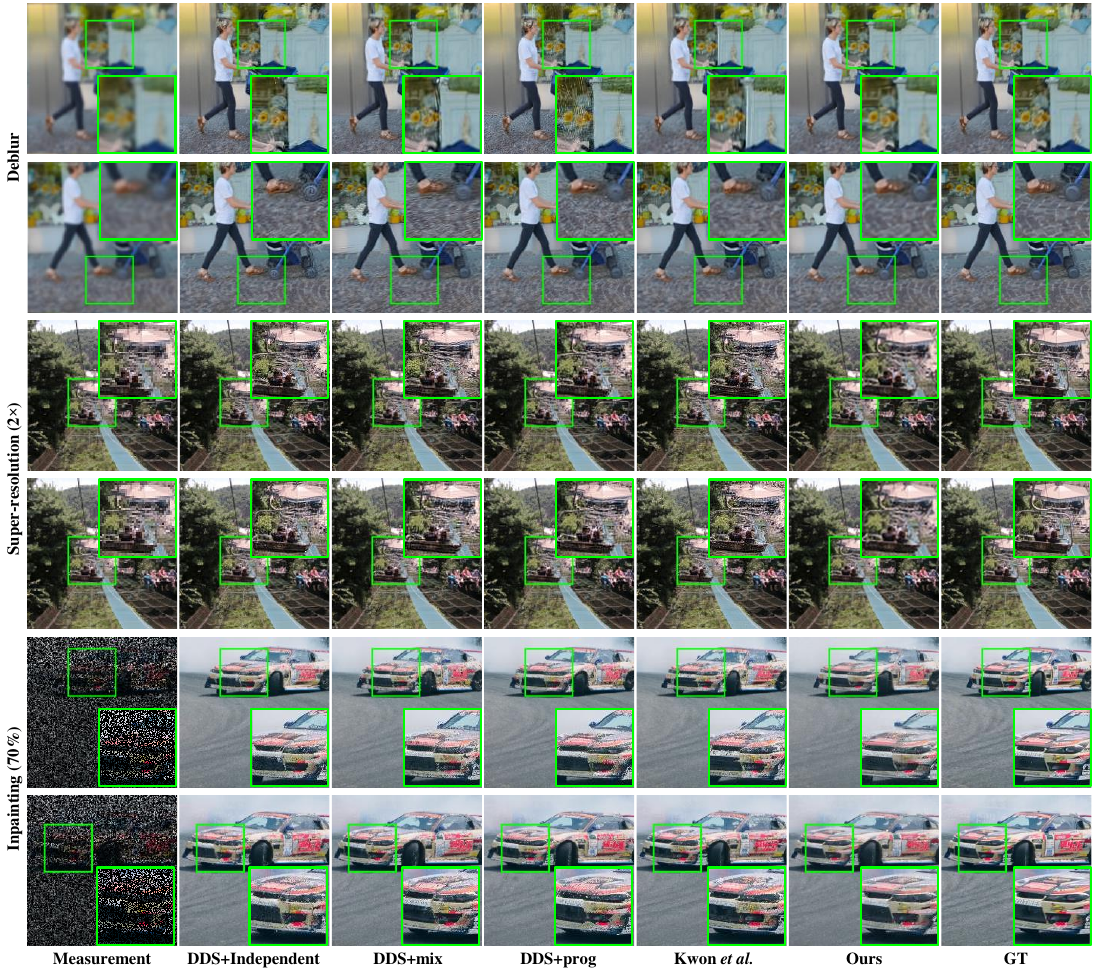}
    \vspace{-2mm}
\caption{Example of various video restoration results including \textit{Gaussian} deblur, super-resolution, and inpainting. Please zoom in for better visual details.}
\vspace{-5mm}
    \label{fig_visual}
\end{figure*}

\begin{table}
    \centering
    \caption{Quantitative evaluation results (PSNR, SSIM, LPIPS, and FVD) of video \textit{Gaussian} deblur on DAVIS~\cite{pont20172017}, REDS4~\cite{nah2019ntire}, and Set8~\cite{tassano2020fastdvdnet}, respectively.
    The best results are marked in \textbf{Bold}.}
    \setlength{\tabcolsep}{0.55mm}
    \adjustbox{width=0.95\linewidth}{
    \begin{tabular}{clcccc}
    \toprule
  
        &Method &PSNR$\uparrow$&SSIM$\uparrow$&LPIPS$\downarrow$&FVD$\downarrow$\\
        \midrule
         \multirow{5}*{\rotatebox{90}{\textbf{DAVIS}~\cite{pont20172017}}}
         &DiffPIR~\cite{zhu2023denoising} + independent&30.51&0.8706&0.1742&37.53\\
         &DiffPIR~\cite{zhu2023denoising} + mix~\cite{ge2023preserve}&30.52&0.8707&0.1741&36.84\\
         &DiffPIR~\cite{zhu2023denoising} + prog~\cite{ge2023preserve}&30.50&0.8705&0.1744&39.74\\
         &DiffPIR~\cite{zhu2023denoising} + identical~\cite{kwon2024solving}&30.52&0.8709&0.1739&38.70\\            &\cellcolor{gray!20}Ours (DiffPIR)&\cellcolor{gray!20}\textbf{30.54}&\cellcolor{gray!20}\textbf{0.8757}&\cellcolor{gray!20}\textbf{0.1621}&\cellcolor{gray!20}\textbf{32.07}\\ 

         \midrule
         \multirow{5}*{\rotatebox{90}{\textbf{REDS4}~\cite{nah2019ntire}}}
         &DiffPIR~\cite{zhu2023denoising} + independent&28.61&0.8662&0.1939&23.83\\
         &DiffPIR~\cite{zhu2023denoising} + mix~\cite{ge2023preserve}&\textbf{28.62}&0.8661&0.1941&24.47\\
         &DiffPIR~\cite{zhu2023denoising} + prog~\cite{ge2023preserve}&28.61&\textbf{0.8663}&0.1939&23.87\\
         &DiffPIR~\cite{zhu2023denoising} + identical~\cite{kwon2024solving}&28.60&0.8661&0.1941&23.21\\         &\cellcolor{gray!20}Ours (DiffPIR)&\cellcolor{gray!20}28.59&\cellcolor{gray!20}0.8653&\cellcolor{gray!20}\textbf{0.1887}&\cellcolor{gray!20}\textbf{22.96}\\ 

        \midrule
         \multirow{5}*{\rotatebox{90}{\textbf{Set8}~\cite{tassano2020fastdvdnet}}}&
         DiffPIR~\cite{zhu2023denoising} + independent&\textbf{30.33}&\textbf{0.8811}&0.1725&33.85\\
         &DiffPIR~\cite{zhu2023denoising} + mix~\cite{ge2023preserve}&30.32&0.8810&0.1725&34.46\\
         &DiffPIR~\cite{zhu2023denoising} + prog~\cite{ge2023preserve}&30.32&0.8809&0.1727&34.21\\
         &DiffPIR~\cite{zhu2023denoising} + identical~\cite{kwon2024solving}&30.32&0.8809&0.1723&33.74\\         &\cellcolor{gray!20}Ours (DiffPIR)&\cellcolor{gray!20}30.17&\cellcolor{gray!20}0.8787&\cellcolor{gray!20}\textbf{0.1616}&\cellcolor{gray!20}\textbf{33.64}\\ 
         
      \bottomrule
    \end{tabular}
    }
    \label{tab_diffpir_exp}
\end{table}

\subsection{Comparison with State-of-the-Art}
To evaluate the performance of our proposed method, we conduct a comparison with several state-of-the-art zero-shot video restoration methods.
Since zero-shot video restoration is an under-explored yet challenging area, there are limited works available to compare~\footnote{\chong{Recent works~\cite{cao2024zero} and~\cite{daras2024warped} do not release the codes for comparison.}}.
Kwon~\etal~\cite{kwon2024solving} is the most related work in this task, where the method adopts an identical Gaussian noise as the initiation and utilizes DDS~\cite{chung2023decomposed} to ensure data consistency during the diffusion sampling process.
To conduct a comprehensive evaluation, we constructed various competing baselines that utilized different noise initial strategies, namely frame-wise independent noise, mix noise model~\cite{ge2023preserve}, progressive noise model~\cite{ge2023preserve}.
To conduct a fair comparison, all of these baselines including our method adopt the same sampling strategy for data consistency, namely DDS~\cite{chung2023decomposed}.
%

\siyuan{Table~\ref{tab_dds_exp} presents the quantitative evaluation results of our proposed method compared to several state-of-the-art approaches, including Gaussian deblur, super-resolution (2× and 4×), and inpainting tasks on the DAVIS, REDS4, and Set8 datasets.}
\siyuan{For Gaussian deblurring, our approach achieves the highest PSNR and SSIM values with lower LPIPS and FVD compared to other methods, indicating better detail preservation and perceptual quality.}
\siyuan{In the super-resolution tasks, our method shows significant improvement for the 2× scaling factor, particularly in PSNR and SSIM. However, for the more challenging 4× super-resolution, some results are less consistent, though our method still maintains competitive perceptual quality.}
\siyuan{For video inpainting, our method achieves superior results in PSNR and SSIM on all datasets.}
\siyuan{Overall, the results in Table 1 demonstrate that our method generally outperforms or is highly competitive with existing zero-shot video restoration techniques, highlighting its robustness and generalization ability.}
To further demonstrate the advantage of our method against other competing baselines, Figure~\ref{fig_visual} provides qualitative comparisons on various video restoration tasks.
Existing methods only adopt a simple noise initiation strategy, leading to significant speckling artifacts across restored frames.
With the merits of the proposed framework, our method consistently aligns closely with the ground truth across all tasks, suggesting robustness that handles various types of image restoration effectively.

\begin{figure*} [t]
  \centering 
    \includegraphics[width=0.99\linewidth]{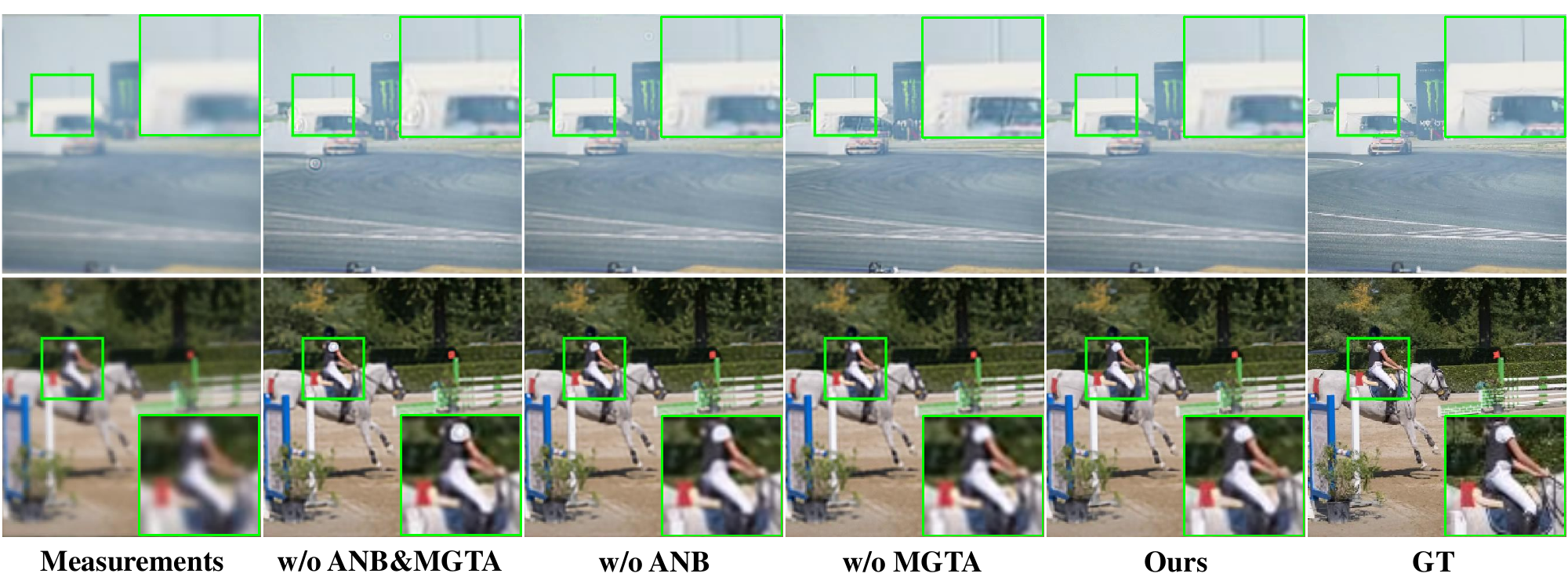}
\caption{Visualization of ablation study by removing each key component of \textit{Gaussian} deblur on the DAVIS dataset.}
    \label{fig_ablation_visual_new}
\end{figure*}

\subsection{Extension to Other Sampling Strategies}
As described in Algorithm~\ref{alg1}, our framework can easily extend existing image-based diffusion sampling strategies to solve video restoration problems with data consistency.
To further demonstrate the generalizability and robustness of the proposed method, we conduct extensive experiments by plug-in DiffPIR~\cite{zhu2023denoising} into our framework on various video datasets.
Following the previous section, we construct competing baseline using DiffPIR combined with frame-wise independent noise, mix noise model~\cite{ge2023preserve}, progressive noise model~\cite{ge2023preserve} and frame-wise identical noise~\cite{kwon2024solving}.
\siyuan{Table~\ref{tab_diffpir_exp} presents the quantitative results using DiffPIR sampling. The proposed method consistently outperforms or remains highly competitive with other approaches across various restoration tasks on all datasets, with particular strength in the LPIPS and FVD.
}


\begin{table}
    \centering
    \caption{Quantitative comparison of the restoration stability.}
    \renewcommand{\arraystretch}{1.4}
    \setlength{\tabcolsep}{0.75mm}
    \adjustbox{width=0.95\linewidth}{
    \begin{tabular}{clccc}
    \toprule
  
        &Method &PSNR$\uparrow$&SSIM$\uparrow$&LPIPS$\downarrow$\\
        \midrule
         \multirow{2}*{\rotatebox{90}{\textbf{DAVIS}}}
         &Kwon~\etal~\cite{kwon2024solving}&27.92$\pm$0.0912&0.8194$\pm$0.0038&0.1875$\pm$0.0017\\            &\cellcolor{gray!20}Ours&\cellcolor{gray!20}\textbf{29.43$\pm$0.0020}&\cellcolor{gray!20}\textbf{0.8608$\pm$0.0001}&\cellcolor{gray!20}\textbf{0.1764$\pm$0.0001}\\ 

         \midrule
         \multirow{2}*{\rotatebox{90}{\textbf{REDS4}}}
         &Kwon~\etal~\cite{kwon2024solving}&25.31$\pm$0.1002&0.7801$\pm$0.0038&0.2211$\pm$0.0014\\         &\cellcolor{gray!20}Ours&\cellcolor{gray!20}\textbf{27.53$\pm$0.0012}&\cellcolor{gray!20}\textbf{0.8357$\pm$0.0001}&\cellcolor{gray!20}\textbf{0.2141$\pm$0.0001}\\ 

        \midrule
         \multirow{2}*{\rotatebox{90}{\textbf{Set8}}}
         &Kwon~\etal~\cite{kwon2024solving}&27.10$\pm$0.1267&0.7957$\pm$0.0056&0.2092$\pm$0.0034\\         &\cellcolor{gray!20}Ours&\cellcolor{gray!20}\textbf{29.10$\pm$0.0013}&\cellcolor{gray!20}\textbf{0.8547$\pm$0.0001}&\cellcolor{gray!20}\textbf{0.1909$\pm$0.0001}\\ 
         
      \bottomrule
    \end{tabular}
    }
    \label{tab_stability}
\end{table}

\subsection{Restoration Stability}
\siyuan{To evaluate the robustness of our proposed method, we analyze the restoration stability of our method compared to competing method~\cite{kwon2024solving} across three datasets. 
As shown in Table~\ref{tab_stability}, our method shows superior stability across all datasets, with lower standard deviations in all metrics.
For example, on the DAVIS dataset, we not only improve the PSNR from 27.92 to 29.43 but also reduce its standard deviation from $\pm$0.0912 to $\pm$0.0020. 
Similar trends are observed for SSIM and LPIPS on the DAVIS dataset, as well as for all three metrics on the REDS4 and Set8 datasets.
These results validate that our noise initialization strategy not only enhances restoration quality but also ensures more reliable and consistent performance.}

\begin{table}
    \centering
    \caption{Ablation study of each proposed component on \textit{Gaussian} deblur on the REDS4~\cite{tassano2020fastdvdnet} dataset.}
    \adjustbox{width=0.95\linewidth}{
    \begin{tabular}{ccc|cccc}
    \toprule
  
        DDIM Inv.&ANB&MGTA &PSNR$\uparrow$&SSIM$\uparrow$&LPIPS$\downarrow$&FVD$\downarrow$\\
        \midrule
        \ding{55}&\ding{55}&\ding{55} &25.29&0.7789&0.2218&59.33\\
        \Checkmark&\ding{55}&\ding{55} &27.07&0.8372&0.2679&71.64\\
        \Checkmark&\ding{55}& \Checkmark&27.47&\textbf{0.8487}&0.2516&82.09\\
        \Checkmark&\Checkmark &\ding{55}&26.98&0.8196&0.2367&93.11\\         \cellcolor{gray!20}\Checkmark&\cellcolor{gray!20}\Checkmark&\cellcolor{gray!20}\Checkmark&\cellcolor{gray!20}\textbf{27.53}&\cellcolor{gray!20}0.8357&\cellcolor{gray!20}\textbf{0.2142}&\cellcolor{gray!20}\textbf{58.85}\\ 
         
      \bottomrule
    \end{tabular}
    }
    \label{tab_ablation_module}
\end{table}

\subsection{Ablation Study}
To investigate the efficacy of each key component of the proposed method, we perform experiments on several variants on the REDS4 dataset.
The ablation comparison starts with a base model employing independent noise initiation alongside DDS sampling.
We incrementally integrated additional components into the framework, specifically DDIM Inversion, Adaptive Noise Blending (ANB), and Measurement-Guided Trajectory Attention (MGTA).
The results in Table~\ref{tab_ablation_module} indicate that using all components together provides the best balance across all metrics, achieving notably low LPIPS and FVD scores, indicative of superior perceptual quality and visual accuracy.
Specifically, utilizing DDIM inversion for noise initiation can effectively improve the PSNR while also introducing unnatural patterns that distort the perceptual quality.
However, with the aid of the proposed modules, such drawbacks can be significantly mitigated, achieving promising performance in both fidelity and visual quality.
Figure~\ref{fig_ablation_visual_new} illustrates a visual comparison on \textit{Gaussian} deblur on the DAIVIS dataset where each key component of our method is removed. 
It clearly demonstrates that without these components, the restoration results suffer from pronounced degradation and noticeable blurring artifacts.

\begin{table}[t]
    \centering
    \caption{
    Comparison of using a fixed Gaussian low-pass filter for noise blending. The best results are marked in \textbf{bold}.}
    \setlength{\tabcolsep}{0.5mm}
    \adjustbox{width=0.85\linewidth}{
    \begin{tabular}{c|ccccc|c}
    \toprule
          \multicolumn{1}{c|}{\ }&\multicolumn{5}{c}{\textbf{Fixed Gaussian Low-pass Filter}}&\multicolumn{1}{|c}{\textbf{Ours}}\\
      \cmidrule(lr){1-1} \cmidrule(lr){2-6} \cmidrule(lr){7-7}
      cut-off freq.&0.1&0.3&0.5&0.7&0.9&adaptive\\
        \midrule
          PSNR$\uparrow$&27.50&29.39&28.81&29.32&29.37&\textbf{29.43}\\ 
          SSIM$\uparrow$&0.8039&0.8400&0.7996&0.8442&0.8567&\textbf{0.8609}\\
          LPIPS$\downarrow$&0.2093&0.1847&0.2022&0.1839&0.1978&\textbf{0.1764}\\
      \bottomrule
    \end{tabular}
    }
    \label{tab_comment1.3a}
\end{table}

\rev{ 
To validate the superiority of our proposed adaptive noise blending strategy, we construct a comparison using a set of fixed Gaussian LPFs with different cut-off frequencies and use them for noise blending.
Concretely, for each frame, we replace the adaptive filter $f(c_i)$ with a frame-invariant $f(c)$, where $c \in \{0.1, 0.3, 0.5, 0.7, 0.9\}$, and keep all other settings (DDIM inversion, sampling steps, and data-consistency updates) unchanged.
Table~\ref{tab_comment1.3a} reports results on video Gaussian deblurring.
We observe that while fixed LPFs can improve over directly using the raw inversion noise, our adaptive filtering achieves the best overall fidelity and perceptual quality (PSNR/SSIM and LPIPS).
This is consistent with our motivation: applying a single pre-defined cut-off frequency to all frames is suboptimal when the frequency distribution varies across frames, since overly aggressive filtering can suppress useful structures for detail-rich frames, whereas overly mild filtering retains more degradation-corrupted components for severely degraded frames.}

\begin{table}[t]
\centering
    \caption{
    Ablation study of the retained energy proportion ($\eta$) in adaptive filtering strategy on the video \textit{Gaussian} deblur on three different datasets DAVIS, REDS4, and Set8. The best results are marked in \textbf{bold}. The \colorbox{gray!20}{gray cells} indicates the selected parameter for the main experiments.}
    \setlength{\tabcolsep}{0.5mm}
    \adjustbox{width=0.99\linewidth}{
    \begin{tabular}{c|ccc|ccc|ccc}
    \toprule
          \multicolumn{1}{c}{\ }&\multicolumn{3}{c}{\textbf{DAVIS}}&\multicolumn{3}{c}{\textbf{REDS4}}&\multicolumn{3}{c}{\textbf{Set8}}\\
      \cmidrule(lr){1-1} \cmidrule(lr){2-4} \cmidrule(lr){5-7} \cmidrule(lr){8-10} 
        $\quad \eta\quad$ &PSNR$\uparrow$&SSIM$\uparrow$&LPIPS$\downarrow$&PSNR$\uparrow$&SSIM$\uparrow$&LPIPS$\downarrow$&PSNR$\uparrow$&SSIM$\uparrow$&LPIPS$\downarrow$\\
        \midrule
         0.1 &27.97&0.8242&0.1995&25.65&0.7868&0.2412&27.46&0.8215&0.2133\\ 
          0.3 &29.04&0.8578&0.1877&26.83&0.8329&0.2308&28.67&0.8583&0.2057\\
          0.5 &\cellcolor{gray!20}29.43&\cellcolor{gray!20}\textbf{0.8607}&\cellcolor{gray!20}\textbf{0.1765}&27.32&\textbf{0.8431}&0.2221&29.02&\textbf{0.8632}&0.1985\\
          0.7 &\textbf{29.48}&0.8467&0.1788&\cellcolor{gray!20}\textbf{27.53}&\cellcolor{gray!20}0.8358&\cellcolor{gray!20}\textbf{0.2139}&\cellcolor{gray!20}\textbf{29.10}&\cellcolor{gray!20}0.8547&\cellcolor{gray!20}\textbf{0.1907}\\
          0.9 &28.85&0.8018&0.2018&27.24&0.7958&0.2352&28.43&0.8092&0.2031\\

      \bottomrule
    \end{tabular}
    }
    \label{tab_comment1.3b}
\end{table}

\rev{
We further analyze the key hyper-parameter $\eta$, which controls the proportion of frequency energy retained by the adaptive filtering.
We evaluate $\eta \in \{0.1, 0.3, 0.5, 0.7, 0.9\}$ on three datasets (DAVIS, REDS4, and Set8) under Gaussian deblurring, using the same experimental settings as the main experiments.
As shown in Table~\ref{tab_comment1.3b}, moderate values of $\eta$ (e.g., $0.5$--$0.7$) generally provide the best performance, while extreme choices lead to degradation:
small $\eta$ tends to over-filter and removes too much informative content from the initialization, making subsequent diffusion sampling harder to recover details, while large $\eta$ retains more high-frequency components from the inversion noise, which may include distorted or degradation-encoded patterns, reducing perceptual quality.
}

To further explore the effectiveness of the proposed measurement-guided trajectory attention mechanism, we conducted additional ablation studies.
Unlike existing methods in video editing~\cite{ceylan2023pix2video, geyer2023tokenflow, qi2023fatezero} which apply spatial-temporal attention during both the inversion and sampling phases, our studies suggest that employing the proposed attention mechanism exclusively during the sampling process enhances restoration performance.
We conduct an ablation study where the DDIM inversion is performed with and without the proposed attention mechanism.
Table~\ref{tab_MGTA} shows the results of the aforementioned two variants on \textit{Gaussian} deblur on three different datasets.
These findings indicate that using the measurement-guided trajectory attention during the inversion process actually hinder the performance, as it reinforces degradation patterns within the frames.
As for video editing, the DDIM inversion is performed on the clean video frames, thus leveraging the spatial-temporal attention will benefit the temporal correlation.
In contrast, in the context of video restoration, the inversion process is performed on the degraded frames.
Consequently, imposing such a temporal-related attention mechanism during the DDIM inverse process will also exploit the correlation across the degradation pattern in each frame, thus hindering restoration performance.

\begin{table}[t]
    \centering
    \caption{Ablation study of incorporating MGTA during the inversion process.}
    \renewcommand{\arraystretch}{1.5}
    \adjustbox{width=0.95\linewidth}{
    \begin{tabular}{clcccc}
    \toprule
  
        &Method &PSNR$\uparrow$&SSIM$\uparrow$&LPIPS$\downarrow$&FVD$\downarrow$\\
        \midrule
         \multirow{2}*{\rotatebox{90}{\textbf{DAVIS}}}
         &Inversion w/ MGTA&29.36 &0.8594 &0.1794 &69.15\\            
         &\cellcolor{gray!20}Inversion w/o MGTA&\cellcolor{gray!20}\textbf{29.43} &\cellcolor{gray!20}\textbf{0.8608}&\cellcolor{gray!20}\textbf{0.1763}&\cellcolor{gray!20}\textbf{63.42}\\ 

         \midrule
         \multirow{2}*{\rotatebox{90}{\textbf{REDS4}}}
         &Inversion w/ MGTA&27.29 &0.8250 &0.2302 &74.05\\         
         &\cellcolor{gray!20}Inversion w/o MGTA&\cellcolor{gray!20}\textbf{27.53} &\cellcolor{gray!20}\textbf{0.8357}&\cellcolor{gray!20}\textbf{0.2142}&\cellcolor{gray!20}\textbf{58.85}\\ 

        \midrule
         \multirow{2}*{\rotatebox{90}{\textbf{Set8}}}
         &Inversion w/ MGTA&28.69 &0.8389 &0.2108 &98.29\\         
         &\cellcolor{gray!20}Inversion w/o MGTA&\cellcolor{gray!20}\textbf{29.10}&\cellcolor{gray!20}\textbf{0.8547}&\cellcolor{gray!20}\textbf{0.1909}&\cellcolor{gray!20}\textbf{81.06}\\ 
         
      \bottomrule
    \end{tabular}
    }
    \vspace{-2mm}
    \label{tab_MGTA}
\end{table}

  


         

\rev{\subsection{Limitations}
Our method improves zero-shot restoration quality and stability through enhanced noise initialization and trajectory-guided temporal attention. 
However, it also introduces additional computational overhead.
Compared with the previous method~\cite{chung2023decomposed}, our current implementation is slower in wall-clock time under the same experimental setting.
The previous method~\cite{chung2023decomposed} takes 9.67~s per video, whereas our full pipeline takes 20.18~s per video.
This increase is mainly due to the proposed components required by our framework, including DDIM inversion for obtaining the initial noise, as well as extra operations such as adaptive noise blending and trajectory attention.
Specifically, the measurement-guided trajectory attention introduces a modest overhead in both runtime and memory, with around +1.94~s runtime and +500~MB memory overhead.
Future work includes accelerating the pipeline by reducing the number of sampling steps, adopting faster initialization strategies, caching or reusing trajectories, and exploring acceleration techniques to obtain a faster feed-forward variant while preserving the proposed stability benefits.}

\section{Conclusion}\label{sec:conclusion}
In this paper, we present a novel zero-shot video restoration framework that only leverages a pre-trained image diffusion model.
We initiate by obtaining correlated noise via DDIM inversion, which can effectively capture and utilize the inherent temporal correlation in the degraded video frames.
Based on that, we propose an adaptive noise blending strategy to remove the degradation pattern encoded during the inversion process while steering the noise toward the Gaussian distribution to enhance the generation ability.
We further propose a measurement-guided trajectory attention to emphasize the temporal consistency during the restoration process.
Finally, extensive experiments on various video restoration tasks demonstrate the superiority of the proposed methods on DAVIS, REDS4 and Set8 datasets.

\bibliographystyle{IEEEtran}
\bibliography{ref}

@inproceedings{zhu2023denoising,
  title={Denoising diffusion models for plug-and-play image restoration},
  author={Zhu, Yuanzhi and Zhang, Kai and Liang, Jingyun and Cao, Jiezhang and Wen, Bihan and Timofte, Radu and Van Gool, Luc},
  booktitle={Proceedings of the IEEE/CVF Conference on Computer Vision and Pattern Recognition},
  pages={1219--1229},
  year={2023}
}

@article{ho2020denoising,
  title={Denoising diffusion probabilistic models},
  author={Ho, Jonathan and Jain, Ajay and Abbeel, Pieter},
  journal={Advances in neural information processing systems},
  volume={33},
  pages={6840--6851},
  year={2020}
}

@inproceedings{sohl2015deep,
  title={Deep unsupervised learning using nonequilibrium thermodynamics},
  author={Sohl-Dickstein, Jascha and Weiss, Eric and Maheswaranathan, Niru and Ganguli, Surya},
  booktitle={International Conference on Machine Learning},
  pages={2256--2265},
  year={2015},
  organization={PMLR}
}

@inproceedings{nichol2021improved,
  title={Improved denoising diffusion probabilistic models},
  author={Nichol, Alexander Quinn and Dhariwal, Prafulla},
  booktitle={International Conference on Machine Learning},
  pages={8162--8171},
  year={2021},
  organization={PMLR}
}

@article{ho2022cascaded,
  title={Cascaded Diffusion Models for High Fidelity Image Generation.},
  author={Ho, Jonathan and Saharia, Chitwan and Chan, William and Fleet, David J and Norouzi, Mohammad and Salimans, Tim},
  journal={J. Mach. Learn. Res.},
  volume={23},
  number={47},
  pages={1--33},
  year={2022}
}

@inproceedings{rombach2022high,
  title={High-resolution image synthesis with latent diffusion models},
  author={Rombach, Robin and Blattmann, Andreas and Lorenz, Dominik and Esser, Patrick and Ommer, Bj{\"o}rn},
  booktitle={Proceedings of the IEEE/CVF Conference on Computer Vision and Pattern Recognition},
  pages={10684--10695},
  year={2022}
}

@article{dhariwal2021diffusion,
  title={Diffusion models beat gans on image synthesis},
  author={Dhariwal, Prafulla and Nichol, Alexander},
  journal={Advances in Neural Information Processing Systems},
  volume={34},
  pages={8780--8794},
  year={2021}
}

@inproceedings{chung2023diffusion,
  title={Diffusion Posterior Sampling for General Noisy Inverse Problems},
  author={Chung, Hyungjin and Kim, Jeongsol and Mccann, Michael T and Klasky, Marc L and Ye, Jong Chul},
  booktitle={The Eleventh International Conference on Learning Representations, ICLR 2023},
  year={2023},
  organization={The International Conference on Learning Representations}
}

@article{chung2022improving,
  title={Improving diffusion models for inverse problems using manifold constraints},
  author={Chung, Hyungjin and Sim, Byeongsu and Ryu, Dohoon and Ye, Jong Chul},
  journal={Advances in Neural Information Processing Systems},
  volume={35},
  pages={25683--25696},
  year={2022}
}

@article{wang2022zero,
  title={Zero-Shot Image Restoration Using Denoising Diffusion Null-Space Model},
  author={Wang, Yinhuai and Yu, Jiwen and Zhang, Jian},
  journal={The Eleventh International Conference on Learning Representations},
  year={2023}
}

@inproceedings{
song2024solving,
title={Solving Inverse Problems with Latent Diffusion Models via Hard Data Consistency},
author={Bowen Song and Soo Min Kwon and Zecheng Zhang and Xinyu Hu and Qing Qu and Liyue Shen},
booktitle={The Twelfth International Conference on Learning Representations},
year={2024},
url={https://openreview.net/forum?id=j8hdRqOUhN}
}

@article{song2020denoising,
  title={Denoising diffusion implicit models},
  author={Song, Jiaming and Meng, Chenlin and Ermon, Stefano},
  journal={arXiv preprint arXiv:2010.02502},
  year={2020}
}

@inproceedings{zhang2018unreasonable,
  title={The unreasonable effectiveness of deep features as a perceptual metric},
  author={Zhang, Richard and Isola, Phillip and Efros, Alexei A and Shechtman, Eli and Wang, Oliver},
  booktitle={Proceedings of the IEEE conference on computer vision and pattern recognition},
  pages={586--595},
  year={2018}
}

@inproceedings{guo2023shadowdiffusion,
  title={Shadowdiffusion: When degradation prior meets diffusion model for shadow removal},
  author={Guo, Lanqing and Wang, Chong and Yang, Wenhan and Huang, Siyu and Wang, Yufei and Pfister, Hanspeter and Wen, Bihan},
  booktitle={Proceedings of the IEEE/CVF Conference on Computer Vision and Pattern Recognition},
  pages={14049--14058},
  year={2023}
}

@inproceedings{guo2023boundary,
  title={Boundary-aware divide and conquer: A diffusion-based solution for unsupervised shadow removal},
  author={Guo, Lanqing and Wang, Chong and Yang, Wenhan and Wang, Yufei and Wen, Bihan},
  booktitle={Proceedings of the IEEE/CVF International Conference on Computer Vision},
  pages={13045--13054},
  year={2023}
}

@inproceedings{gao2023implicit,
  title={Implicit diffusion models for continuous super-resolution},
  author={Gao, Sicheng and Liu, Xuhui and Zeng, Bohan and Xu, Sheng and Li, Yanjing and Luo, Xiaoyan and Liu, Jianzhuang and Zhen, Xiantong and Zhang, Baochang},
  booktitle={Proceedings of the IEEE/CVF conference on computer vision and pattern recognition},
  pages={10021--10030},
  year={2023}
}

@inproceedings{ren2023multiscale,
  title={Multiscale structure guided diffusion for image deblurring},
  author={Ren, Mengwei and Delbracio, Mauricio and Talebi, Hossein and Gerig, Guido and Milanfar, Peyman},
  booktitle={Proceedings of the IEEE/CVF International Conference on Computer Vision},
  pages={10721--10733},
  year={2023}
}

@inproceedings{lugmayr2022repaint,
  title={Repaint: Inpainting using denoising diffusion probabilistic models},
  author={Lugmayr, Andreas and Danelljan, Martin and Romero, Andres and Yu, Fisher and Timofte, Radu and Van Gool, Luc},
  booktitle={Proceedings of the IEEE/CVF conference on computer vision and pattern recognition},
  pages={11461--11471},
  year={2022}
}

@inproceedings{zhou2024upscale,
  title={Upscale-A-Video: Temporal-Consistent Diffusion Model for Real-World Video Super-Resolution},
  author={Zhou, Shangchen and Yang, Peiqing and Wang, Jianyi and Luo, Yihang and Loy, Chen Change},
  booktitle={Proceedings of the IEEE/CVF Conference on Computer Vision and Pattern Recognition},
  pages={2535--2545},
  year={2024}
}

@inproceedings{yang2025motion,
  title={Motion-guided latent diffusion for temporally consistent real-world video super-resolution},
  author={Yang, Xi and He, Chenhang and Ma, Jianqi and Zhang, Lei},
  booktitle={European Conference on Computer Vision},
  pages={224--242},
  year={2025},
  organization={Springer}
}

@article{kwon2024solving,
  title={Solving Video Inverse Problems Using Image Diffusion Models},
  author={Kwon, Taesung and Ye, Jong Chul},
  journal={arXiv preprint arXiv:2409.02574},
  year={2024}
}

@article{cao2024zero,
  title={Zero-shot Video Restoration and Enhancement Using Pre-Trained Image Diffusion Model},
  author={Cao, Cong and Yue, Huanjing and Liu, Xin and Yang, Jingyu},
  journal={arXiv preprint arXiv:2407.01960},
  year={2024}
}

@inproceedings{ceylan2023pix2video,
  title={Pix2video: Video editing using image diffusion},
  author={Ceylan, Duygu and Huang, Chun-Hao P and Mitra, Niloy J},
  booktitle={Proceedings of the IEEE/CVF International Conference on Computer Vision},
  pages={23206--23217},
  year={2023}
}

@inproceedings{qi2023fatezero,
  title={Fatezero: Fusing attentions for zero-shot text-based video editing},
  author={Qi, Chenyang and Cun, Xiaodong and Zhang, Yong and Lei, Chenyang and Wang, Xintao and Shan, Ying and Chen, Qifeng},
  booktitle={Proceedings of the IEEE/CVF International Conference on Computer Vision},
  pages={15932--15942},
  year={2023}
}

@article{geyer2023tokenflow,
  title={Tokenflow: Consistent diffusion features for consistent video editing},
  author={Geyer, Michal and Bar-Tal, Omer and Bagon, Shai and Dekel, Tali},
  journal={arXiv preprint arXiv:2307.10373},
  year={2023}
}

@article{wang2024cove,
  title={COVE: Unleashing the Diffusion Feature Correspondence for Consistent Video Editing},
  author={Wang, Jiangshan and Ma, Yue and Guo, Jiayi and Xiao, Yicheng and Huang, Gao and Li, Xiu},
  journal={arXiv preprint arXiv:2406.08850},
  year={2024}
}

@inproceedings{wu2023tune,
  title={Tune-a-video: One-shot tuning of image diffusion models for text-to-video generation},
  author={Wu, Jay Zhangjie and Ge, Yixiao and Wang, Xintao and Lei, Stan Weixian and Gu, Yuchao and Shi, Yufei and Hsu, Wynne and Shan, Ying and Qie, Xiaohu and Shou, Mike Zheng},
  booktitle={Proceedings of the IEEE/CVF International Conference on Computer Vision},
  pages={7623--7633},
  year={2023}
}

@article{chung2023decomposed,
  title={Decomposed Diffusion Sampler for Accelerating Large-Scale Inverse Problems},
  author={Chung, Hyungjin and Lee, Suhyeon and Ye, Jong Chul},
  journal={arXiv preprint arXiv:2303.05754},
  year={2023}
}

@article{pont20172017,
  title={The 2017 davis challenge on video object segmentation},
  author={Pont-Tuset, Jordi and Perazzi, Federico and Caelles, Sergi and Arbel{\'a}ez, Pablo and Sorkine-Hornung, Alex and Van Gool, Luc},
  journal={arXiv preprint arXiv:1704.00675},
  year={2017}
}

@inproceedings{nah2019ntire,
  title={Ntire 2019 challenge on video deblurring and super-resolution: Dataset and study},
  author={Nah, Seungjun and Baik, Sungyong and Hong, Seokil and Moon, Gyeongsik and Son, Sanghyun and Timofte, Radu and Mu Lee, Kyoung},
  booktitle={Proceedings of the IEEE/CVF conference on computer vision and pattern recognition workshops},
  pages={0--0},
  year={2019}
}

@inproceedings{tassano2020fastdvdnet,
  title={Fastdvdnet: Towards real-time deep video denoising without flow estimation},
  author={Tassano, Matias and Delon, Julie and Veit, Thomas},
  booktitle={Proceedings of the IEEE/CVF conference on computer vision and pattern recognition},
  pages={1354--1363},
  year={2020}
}

@article{unterthiner2019fvd,
  title={FVD: A new metric for video generation},
  author={Unterthiner, Thomas and van Steenkiste, Sjoerd and Kurach, Karol and Marinier, Rapha{\"e}l and Michalski, Marcin and Gelly, Sylvain},
  year={2019}
}

@inproceedings{garber2024image,
  title={Image restoration by denoising diffusion models with iteratively preconditioned guidance},
  author={Garber, Tomer and Tirer, Tom},
  booktitle={Proceedings of the IEEE/CVF Conference on Computer Vision and Pattern Recognition},
  pages={25245--25254},
  year={2024}
}

@inproceedings{chen2024learning,
  title={Learning Spatial Adaptation and Temporal Coherence in Diffusion Models for Video Super-Resolution},
  author={Chen, Zhikai and Long, Fuchen and Qiu, Zhaofan and Yao, Ting and Zhou, Wengang and Luo, Jiebo and Mei, Tao},
  booktitle={Proceedings of the IEEE/CVF Conference on Computer Vision and Pattern Recognition},
  pages={9232--9241},
  year={2024}
}

@article{yeh2024diffir2vr,
  title={DiffIR2VR-Zero: Zero-Shot Video Restoration with Diffusion-based Image Restoration Models},
  author={Yeh, Chang-Han and Lin, Chin-Yang and Wang, Zhixiang and Hsiao, Chi-Wei and Chen, Ting-Hsuan and Shiu, Hau-Shiang and Liu, Yu-Lun},
  journal={arXiv preprint arXiv:2407.01519},
  year={2024}
}

@inproceedings{ge2023preserve,
  title={Preserve your own correlation: A noise prior for video diffusion models},
  author={Ge, Songwei and Nah, Seungjun and Liu, Guilin and Poon, Tyler and Tao, Andrew and Catanzaro, Bryan and Jacobs, David and Huang, Jia-Bin and Liu, Ming-Yu and Balaji, Yogesh},
  booktitle={Proceedings of the IEEE/CVF International Conference on Computer Vision},
  pages={22930--22941},
  year={2023}
}

@article{ho2022video,
  title={Video diffusion models},
  author={Ho, Jonathan and Salimans, Tim and Gritsenko, Alexey and Chan, William and Norouzi, Mohammad and Fleet, David J},
  journal={Advances in Neural Information Processing Systems},
  volume={35},
  pages={8633--8646},
  year={2022}
}

@article{harvey2022flexible,
  title={Flexible diffusion modeling of long videos},
  author={Harvey, William and Naderiparizi, Saeid and Masrani, Vaden and Weilbach, Christian and Wood, Frank},
  journal={Advances in Neural Information Processing Systems},
  volume={35},
  pages={27953--27965},
  year={2022}
}

@article{xing2023survey,
  title={A survey on video diffusion models},
  author={Xing, Zhen and Feng, Qijun and Chen, Haoran and Dai, Qi and Hu, Han and Xu, Hang and Wu, Zuxuan and Jiang, Yu-Gang},
  journal={ACM Computing Surveys},
  year={2023},
  publisher={ACM New York, NY}
}

@article{ho2022imagen,
  title={Imagen video: High definition video generation with diffusion models},
  author={Ho, Jonathan and Chan, William and Saharia, Chitwan and Whang, Jay and Gao, Ruiqi and Gritsenko, Alexey and Kingma, Diederik P and Poole, Ben and Norouzi, Mohammad and Fleet, David J and others},
  journal={arXiv preprint arXiv:2210.02303},
  year={2022}
}

@inproceedings{wang2019edvr,
  title={Edvr: Video restoration with enhanced deformable convolutional networks},
  author={Wang, Xintao and Chan, Kelvin CK and Yu, Ke and Dong, Chao and Change Loy, Chen},
  booktitle={Proceedings of the IEEE/CVF conference on computer vision and pattern recognition workshops},
  pages={0--0},
  year={2019}
}

@inproceedings{zhou2022revisiting,
  title={Revisiting temporal alignment for video restoration},
  author={Zhou, Kun and Li, Wenbo and Lu, Liying and Han, Xiaoguang and Lu, Jiangbo},
  booktitle={Proceedings of the IEEE/CVF conference on computer vision and pattern recognition},
  pages={6053--6062},
  year={2022}
}

@inproceedings{kim2018spatio,
  title={Spatio-temporal transformer network for video restoration},
  author={Kim, Tae Hyun and Sajjadi, Mehdi SM and Hirsch, Michael and Scholkopf, Bernhard},
  booktitle={Proceedings of the European conference on computer vision (ECCV)},
  pages={106--122},
  year={2018}
}

@article{cheng2021audio,
  title={Audio-driven talking video frame restoration},
  author={Cheng, Harry and Guo, Yangyang and Yin, Jianhua and Chen, Haonan and Wang, Jiafang and Nie, Liqiang},
  journal={IEEE Transactions on Multimedia},
  year={2021},
  publisher={IEEE}
}

@article{chen2023self,
  title={Self-supervised video representation learning by serial restoration with elastic complexity},
  author={Chen, Ziyu and Wang, Hanli and Chen, Chang Wen},
  journal={IEEE Transactions on Multimedia},
  year={2023},
  publisher={IEEE}
}

@article{ding2023blind,
  title={Blind Quality Enhancement for Compressed Video},
  author={Ding, Qing and Shen, Liquan and Yu, Liangwei and Yang, Hao and Xu, Mai},
  journal={IEEE Transactions on Multimedia},
  year={2023},
  publisher={IEEE}
}

@article{baniya2023omnidirectional,
  title={Omnidirectional video super-resolution using deep learning},
  author={Baniya, Arbind Agrahari and Lee, Tsz-Kwan and Eklund, Peter W and Aryal, Sunil},
  journal={IEEE Transactions on Multimedia},
  volume={26},
  pages={540--554},
  year={2023},
  publisher={IEEE}
}

@article{jiang2024animediff,
  title={AnimeDiff: Customized Image Generation of Anime Characters Using Diffusion Model},
  author={Jiang, Yuqi and Liu, Qiankun and Chen, Dongdong and Yuan, Lu and Fu, Ying},
  journal={IEEE Transactions on Multimedia},
  year={2024},
  publisher={IEEE}
}

@article{zhang2024mmginpainting,
  title={MMGInpainting: Multi-Modality Guided Image Inpainting Based On Diffusion Models},
  author={Zhang, Cong and Yang, Wenxia and Li, Xin and Han, Huan},
  journal={IEEE Transactions on Multimedia},
  year={2024},
  publisher={IEEE}
}

@article{xu2024sgdm,
  title={SGDM: An Adaptive Style-Guided Diffusion Model for Personalized Text to Image Generation},
  author={Xu, Yifei and Xu, Xiaolong and Gao, Honghao and Xiao, Fu},
  journal={IEEE Transactions on Multimedia},
  year={2024},
  publisher={IEEE}
}

@article{daras2024warped,
  title={Warped diffusion: Solving video inverse problems with image diffusion models},
  author={Daras, Giannis and Nie, Weili and Kreis, Karsten and Dimakis, Alex and Mardani, Morteza and Kovachki, Nikola and Vahdat, Arash},
  journal={Advances in Neural Information Processing Systems},
  volume={37},
  pages={101116--101143},
  year={2024}
}

@article{liang2024vrt,
  title={Vrt: A video restoration transformer},
  author={Liang, Jingyun and Cao, Jiezhang and Fan, Yuchen and Zhang, Kai and Ranjan, Rakesh and Li, Yawei and Timofte, Radu and Van Gool, Luc},
  journal={IEEE Transactions on Image Processing},
  volume={33},
  pages={2171--2182},
  year={2024},
  publisher={IEEE}
}

@article{liang2022recurrent,
  title={Recurrent video restoration transformer with guided deformable attention},
  author={Liang, Jingyun and Fan, Yuchen and Xiang, Xiaoyu and Ranjan, Rakesh and Ilg, Eddy and Green, Simon and Cao, Jiezhang and Zhang, Kai and Timofte, Radu and Gool, Luc V},
  journal={Advances in Neural Information Processing Systems},
  volume={35},
  pages={378--393},
  year={2022}
}

\vfill

\end{document}